\documentclass[10pt,journal,compsoc]{IEEEtran}

\usepackage{graphicx}

\usepackage{float}
\usepackage{graphicx}
\usepackage{comment}
\usepackage{amsmath,amssymb} 
\usepackage{color}
\usepackage{xspace}
\usepackage{textcomp}

\usepackage{mathtools}
\usepackage{bbm}
\usepackage{makecell, verbatim, sidecap}
\usepackage{subfig}
\usepackage{multirow}
\usepackage{url,nicefrac}

\usepackage[nocompress]{cite}
\usepackage[margin=7pt,font=small,labelfont=bf,labelsep=endash,tableposition=bottom]{caption}

\usepackage{threeparttable}
\usepackage[vlined, ruled, linesnumbered]{algorithm2e}
\usepackage{multirow}
\usepackage{booktabs}
\usepackage{bbm}
\usepackage{pifont}
\usepackage{subfig}
\usepackage{graphicx}
\usepackage{amsmath}
\usepackage{enumitem}
\usepackage{eucal,nicefrac}
\usepackage{bm}
\usepackage{mathrsfs,xspace,xfrac}  

\usepackage{xcolor,colortbl}

\renewcommand{\mathscr}[1]{ {{\mathcal #1} } }

\renewcommand{\vec}[1]{\ensuremath{\pmb{#1}}}
\newcommand{\mat}[1]{\ensuremath{\mathbf{#1}}}
\newcommand{\set}[1]{\ensuremath{\mathscr{#1}}}

\makeatletter
\@tfor\next:=abcdefghijklmnopqrstuvwxyxz\do
{\begingroup\edef\x{\endgroup
		\noexpand\@namedef{v\next}{\noexpand\vec{\next}}%
	}\x}
\@tfor\next:=ABCDEFGHIJKLMNOPQRSTUVWXYZ\do
{\begingroup\edef\x{\endgroup
		\noexpand\@namedef{m\next}{\noexpand\mat{\next}}%
	}\x}
\@tfor\next:=ABCDEFGHIJKLMNOPQRSTUVWXYZ\do
{\begingroup\edef\x{\endgroup
		\noexpand\@namedef{s\next}{\noexpand\set{\next}}%
	}\x}
\makeatother

\def\eg{{\it e.g.}\xspace}
\def\ie{{\it i.e.}\xspace}

\newcommand{\etal}{\textit{et al.}\xspace}

\usepackage{xcolor}

\begin{document}

\title{Towards Accurate Reconstruction of 3D Scene Shape from A Single Monocular Image}

\author{Wei Yin, ~ Jianming Zhang, ~ Oliver Wang, ~
        Simon Niklaus,~ Simon Chen, ~ Yifan Liu,
        ~ Chunhua Shen
\IEEEcompsocitemizethanks{\IEEEcompsocthanksitem 
WY and YL are with The University of Adelaide, Australia.
CS is with Zhejiang University, China.
Part of this work was done when 
CS was with The University of Adelaide, Australia.
JZ, OW, SN, and SC are with Adobe.
\IEEEcompsocthanksitem 
CS is the corresponding author
(e-mail: chunhua@me.com).
}%
\thanks{
    \today{}.
}
}
    
\IEEEtitleabstractindextext{
\begin{abstract}
    Despite significant progress made in the past few years, 
    challenges remain for depth estimation using a single monocular image. First,  it is nontrivial to train a  metric-depth 
    prediction model that can generalize well %
    to 
    diverse scenes
    mainly due to limited training data. 
    Thus, researchers have  built large-scale relative depth datasets 
    that are much easier to collect. 
    However, existing relative depth estimation models often fail to 
    recover accurate 3D scene shapes due to the unknown depth shift 
    caused by training with the relative depth data. 
    \iffalse
    %
    Although adding sparse guided points as input may make it easier to obtain metric depth, the generalization to different sparsity patterns and diverse scenes remains challenging.
    \fi 
    %
    %
We %
tackle this problem here and attempt to estimate accurate scene shapes 
by training on large-scale relative depth data, and estimating the depth shift. 
To do so, we propose a two-stage framework that first predicts depth up to an unknown scale and shift from a single monocular image, and then 
exploits 3D point cloud data %
to predict the %
depth shift and the camera's focal length that allow us to recover %
3D scene shapes. 
As the two modules are trained separately, we do not  need %
strictly paired training data. In addition, we propose an image-level normalized regression loss and a normal-based geometry loss to %
improve %
training with relative depth annotation. 
We test our depth model on nine unseen datasets and achieve state-of-the-art performance
on zero-shot evaluation.
Code is available at:
{   
    \def\UrlFont{\small\sf  \color{blue}} 
    \url{https://github.com/aim-uofa/depth/}
}

\end{abstract}


}

\maketitle

\begin{figure*}[h!]
    \centering
    \includegraphics[width=0.7485\textwidth]{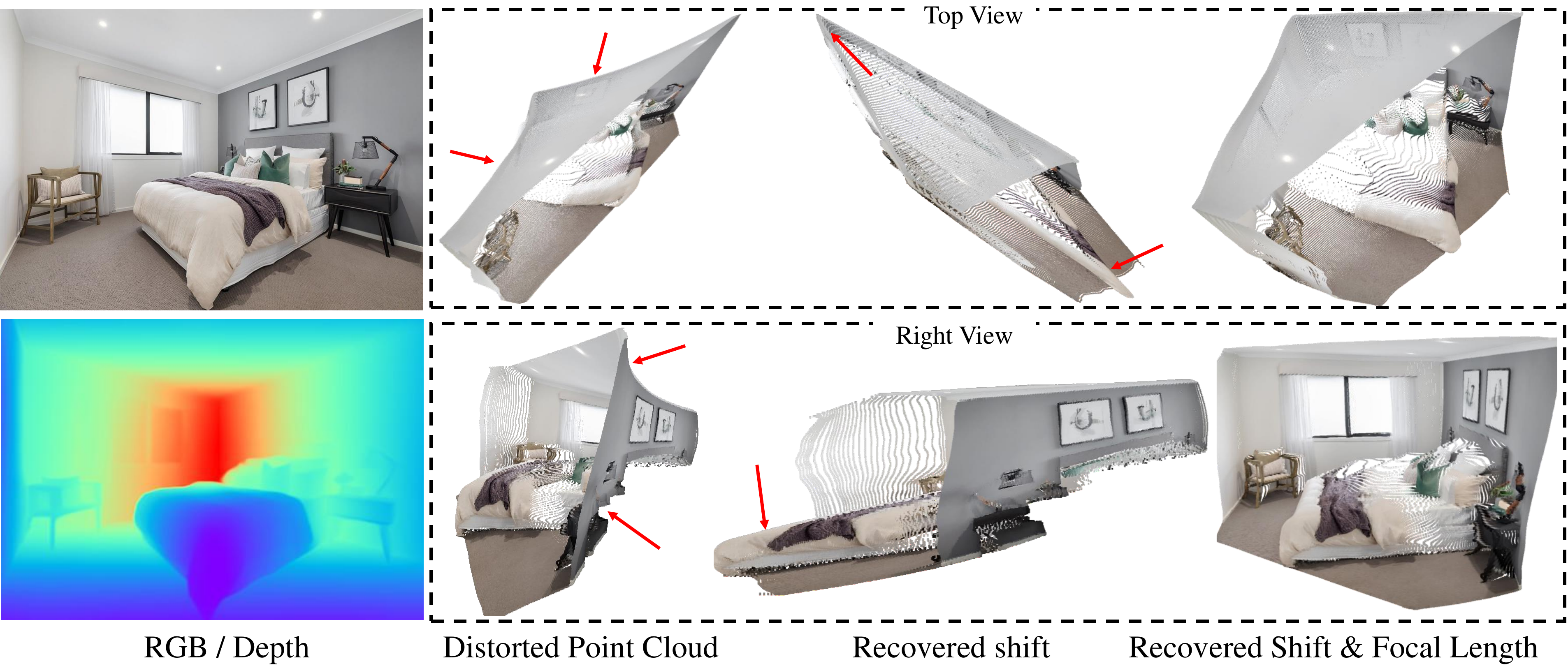}
    \caption
    {
    \textbf
    {
        3D scene structure distortion of projected point clouds.
    } 
    While the predicted depth map appears very  %
    good, 
    the 3D scene shape of the point cloud suffers from noticeable distortions due to an unknown depth shift and focal length (%
2nd column). Our method recovers these parameters using the 3D point cloud %
    information. 
    With the recovered depth shift, the wall and bed edges become straight. However,  
    the overall scene is     
    stretched (3rd column). Finally, with recovered focal length, an accurate 3D scene can be reconstructed (4th column).
}    
    \label{Fig: first page fig.}
    \vspace{-1 em}
\end{figure*}

\section{Introduction}
3D scene reconstruction is a fundamental task in computer vision.
Current established approaches to address this task mainly employ  %
multi-view geometry \cite{hartley2003multiple}, which reconstructs 3D 
scenes based on feature-point correspondence with consecutive frames or multiple views. 
In contrast, %
we 
aim to recover \textit{ 
dense 3D scene shape up to a scale from a single in-the-wild image. With sparse guided depth points, our method can further achieve metric shape.
    }
From a single image input, some methods~\cite{denninger2020} propose to  reconstruct both seen and occlusion surfaces and represent them in a volumetric model or 3D meshes, while our method only recovers the seen surfaces and use the point cloud for representation.
Under this setting, without multiple views available, we rely on monocular depth estimation. 
However, as shown in Fig.~\ref{Fig: first page fig.}, existing monocular depth estimation methods
\cite{xian2018monocular,liu2015learning, Yin2019enforcing} alone are unable to faithfully recover an accurate 3D point cloud. Even with sparse guided points, it is still challenging to generalize to diverse scenes. The key challenges are:
1) it is difficult to collect large-scale metric depth datasets with 
diverse scenes, 
which are needed to achieve good monocular depth estimation models;
2) alternatively, one can train models on large-scale \textit{relative}
depth datasets which are much easier to collect. 
We discover that 
learning depth on such datasets requires estimating the depth shift and focal length to generate accurate 3D scene shapes. 
This problem was %
rarely 
studied in the literature, %
which 
we attempt to tackle %
here.

Recent works have shown %
great progress 
by training deep neural networks on diverse in-the-wild data, \eg,  web stereo images and stereo videos  \cite{chen2016single, chen2020oasis, Ranftl2020, wang2019web,  xian2018monocular, xian2020structure, yin2020diversedepth}. 
Chen \etal\  \cite{chen2016single} propose the first large-scale and in-the-wild dataset, termed DIW. 
Each image only provides a pair of points and annotates their depth relations, \ie, one is farther or closer than the other one. %
Xian \etal\ \cite{xian2018monocular} propose to collect diverse web stereo images and use optical flow %
for finding pixel-wise correspondence so as to create dense \textit{relative} ground-truth depth because camera parameters are unknown and differ for each pair of stereo images.

However, 
web stereo images and videos can only provide depth supervision up to a scale and shift due to the unknown camera baselines and stereoscopic post-processing \cite{lang2010nonlinear}. 
Moreover, 
the diversity of the training data also poses challenges for the model training, as training data captured by different cameras can exhibit significantly different image priors for depth estimation \cite{facil2019cam}.

As a result, state-of-the-art in-the-wild monocular depth estimation models use various types of objective functions that are 
invariant to scale and shift to facilitate training. 
While an unknown scale in depth %
does 
not cause %
scene shape distortion, as it scales the 3D scene shape uniformly, an unknown depth shift %
does 
(see Sec.~\ref{sec:3D_shape}. As shown in Fig.~\ref{Fig: first page fig.}, 
the walls are not flat because of the unknown shift). 
In addition, the camera focal length of a given image may not be accessible at the testing time, leading to more distortion of the 3D scene shape (see the angle between two walls of ``Recovered shift''  in Fig.~\ref{Fig: first page fig.}).  %
This scene shape distortion 
can cause 
critical problems  for some downstream tasks such as  3D view synthesis and 3D photography.

To address these challenges, we propose a novel two-stage monocular scene shape estimation framework that consists of 1) a depth prediction module;
and 2) a point cloud reconstruction module. 
The depth prediction module is a convolutional network trained on a mix of existing datasets that predicts depth maps up to a scale and shift. 
The point cloud reconstruction module leverages point cloud encoder networks that predict shift and focal length adjustment factors from an initial guess of the scene point cloud reconstruction. 
A key observation 
that 
we make here is that,  \textit{when operating on point clouds derived from depth maps, and not on images themselves, we can train models to learn 3D scene shapes %
using synthetic 3D data or data acquired by 3D laser scanning devices.
The domain gap is
significantly less of an issue for point clouds than that for images. 
}%
We empirically show that the  point cloud %
network 
generalizes well to unseen datasets. Moreover, as two modules can be trained separately,
we do not need %
paired ``RGB-Point Cloud'' training data.

To obtain a robust model, we propose to mix multiple sources of data for training, including high-quality   LiDAR sensor data, medium-quality calibrated stereo data, and low-quality web stereo data. Considering %
the quality difference, 
we propose to distinguish them and %
use 
heterogeneous losses instead of a uniform form. For example, the low-quality data can only provide reliable depth ordinal relations, thus the ranking loss \cite{xian2018monocular, xian2020structure} is 
used. 
Other data sources may have more accurate %
depths, but cameras are various.
The training schedule on multiple heterogeneous data sources 
can %
show great 
impact on the final performance.
We propose a simple %
yet 
effective %
normalized regression loss for high-quality and medium-quality data. It transforms the depth data to a canonical scale-shift-invariant space for more robust training. Furthermore,
to improve the geometry quality of the depth, we propose a pair-wise normal regression loss, which can account for both local and global geometry constraints. From high-quality data, reliable local normal information and global planes relations %
are 
extracted, while other data can only provide co-plane information from semantics. Explicitly %
using 
these relations can significantly improve the depth quality.

From a single image input, we can reconstruct the scene shape up to a scale. To further retrieve metric reconstruction, we input some sparse depth points as guidance. However, current state-of-the-art depth completion methods~\cite{park2019deepsdf, cheng2020cspn++} often suffer from %
poor 
generalization.  
They work well on one specific sparsity pattern but generalize poorly to other types of sparse depth. 
They are often sensitive to the domain source of the data and the noisy sparse inputs.
To solve these limitations, first, we employ our proposed heterogeneous losses for training on mixed datasets. Second,  we create  a  diverse  set  of  synthetic  sparsity  patterns  in training to improve the model's robustness to various sparsity types. Furthermore,  to make our method robust to noise, we leverage the depth map predicted by our single image depth prediction method as a data-driven scene prior. By incorporating sparse metric depth cues and a single image relative
depth prior, our method is able to robustly produce a metric depth map.

To summarize, our main contributions are as follows.
\begin{itemize}[noitemsep]
    
    \item We propose a novel framework for in-the-wild monocular 3D scene shape estimation. 
    To 
    our knowledge, this is the first %
    approach tackling 
    this task, and the first method to leverage 3D point cloud neural networks for improving the estimation of the structure of point clouds derived from depth maps.
    
    \item We propose an image-wise normalized regression loss and a pair-wise normal regression loss for improving monocular depth estimation models trained on mixed multi-source datasets.

    {\item 
    We propose to employ the mixed data training strategy for %
    depth completion, which can boost its robustness to diverse scenes, various sparsity patterns, and noisy inputs. %
    Remarkably, 
    With very sparse depth points and even noisy inputs, we 
    show 
    promising 
    metric reconstruction results.}
    
\end{itemize}
Experiments show that our point cloud reconstruction %
method 
can recover accurate 3D shapes from single monocular images (up to scale). Moreover, with sparse points as guidance, we can obtain metric 3D shapes.
Also, 
for depth prediction, %
our method 
achieves state-of-the-art results on zero-shot %
evaluation on 
$10$ unseen datasets. Our depth completion is much more robust than state-of-the-art methods %
on 
diverse scenes and various sparsity types.

\subsection{Related Work}
\noindent\textbf{Monocular depth estimation.} 
Monocular depth prediction is an ill-posed problem by nature.
Many supervised and self-supervised methods  \cite{monodepth2, FuCVPR18-DORN, bian2019unsupervised, eigen2014depth, yin2018geonet, Yin2019enforcing} have been proposed to improve the performance on %
benchmarks, such as NYU  \cite{silberman2012indoor} and KITTI  \cite{geiger2012we}. Fu~\etal  \cite{FuCVPR18-DORN} propose to apply the atrous spatial pyramid pooling (ASPP), 
and enforce the ordinal regression loss to predict the metric depth.
Yin~\etal  \cite{yin2018geonet} propose the high-order virtual normal loss to leverage the %
long-range 
geometry structure, %
which %
significantly improves the quality of the reconstructed 3D point cloud.  
Godard~\etal  \cite{monodepth2} propose a minimum re-projection loss and an auto-masking loss to leverage the geometry relations between consecutive frames and left-right views under the self-supervised learning framework. It can achieve comparable performance on KITTI with state-of-the-art supervised learning methods. Although impressive performance can be achieved, %
all of these
metric depth prediction  methods only work well on limited scenes, and %
poorly generalize to diverse scenes.

To solve this problem, %
a few
methods are proposed to tackle the issue of estimating depth ``in the wild''. %
Recently, we have witnessed %
impressive progress \cite{chen2016single, chen2020oasis, wang2019web,  wang2020foresee, xian2018monocular, xian2020structure, yin2020diversedepth}. 
The key here %
is to collect large-scale training data of either metric depth or relative depth. 
When metric depth supervision is available, networks can be trained to directly regress the metric depth  \cite{eigen2014depth, liu2015learning, Yin2019enforcing}. 
However, obtaining metric ground-truth depth for diverse datasets is challenging. 
On the other hand, 
Chen~\etal  \cite{chen2016single} collect the first diverse \emph{relative} depth annotations for internet images, while other approaches propose to %
employ 
stereo images or videos from the internet  \cite{Ranftl2020, wang2019web,  xian2018monocular, xian2020structure, yin2020diversedepth}.
Such diverse data is important for      good    generalization.
As the metric depth is unavailable, depth regression losses cannot be used directly. 
Instead, these methods rely either on the ranking losses 
\cite{chen2016single, xian2018monocular, xian2020structure} or scale and shift invariant loss functions \cite{Ranftl2020, wang2019web}.
In general, the ranking loss alone cannot fully exploit the supervision and thus can only produce coarse depth maps.
As the camera model is unknown, thus 
the 
depth can be predicted only up to scale and shift 
(so-called affine-invariant depth estimation~\cite{yin2020diversedepth, yin2021virtual}). 
The 3D shape cannot be reconstructed from the predicted depth maps accurately. Here, we tackle this problem:  %
we aim to accurately recover the 3D shape from a single image (\textit{i.e.},  \textit{3D scene reconstruction up to a scale}). 

\noindent\textbf{3D reconstruction from a single image.} 
Several works have addressed reconstructing %
various 
types of objects from a single image  \cite{wang2018pixel2mesh, wu2018learning, Xu2022Towards3S}. 
According to the representation used, methods can be classified %
into 
voxels \cite{wu2016learning, tatarchenko2019single}, meshes \cite{wang2018pixel2mesh, ranjan2018generating}, point clouds, and implicit functions. 
Both rigid and non-rigid object reconstruction was studied in the literature.
For example, 
Pixel2Mesh \cite{wang2018pixel2mesh} proposes to reconstruct the 3D shape from a single image and express it in a triangular mesh. Recently, 
continuous implicit
functions \cite{mescheder2019occupancy, park2019deepsdf} %
are employed 
to represent shapes \cite{saito2019pifu, saito2020pifuhd}. 
Note that, these methods cannot be directly applied to reconstruct 3D scenes from single images.

On the other hand, %
a few methods reconstruct 
3D scenes %
from a single image.
Saxena \etal \cite{saxena2008make3d}
segment a scene into planes and 
predict the orientation and the location of these planes and stitch them together to represent the scene. 
Other works propose to use image cues, such as shading  \cite{prados2005shape} and contour edges  \cite{karpenko2006smoothsketch} for the scene reconstruction. 

\begin{figure*}[t]
\centering
\includegraphics[width=.83690751\textwidth]{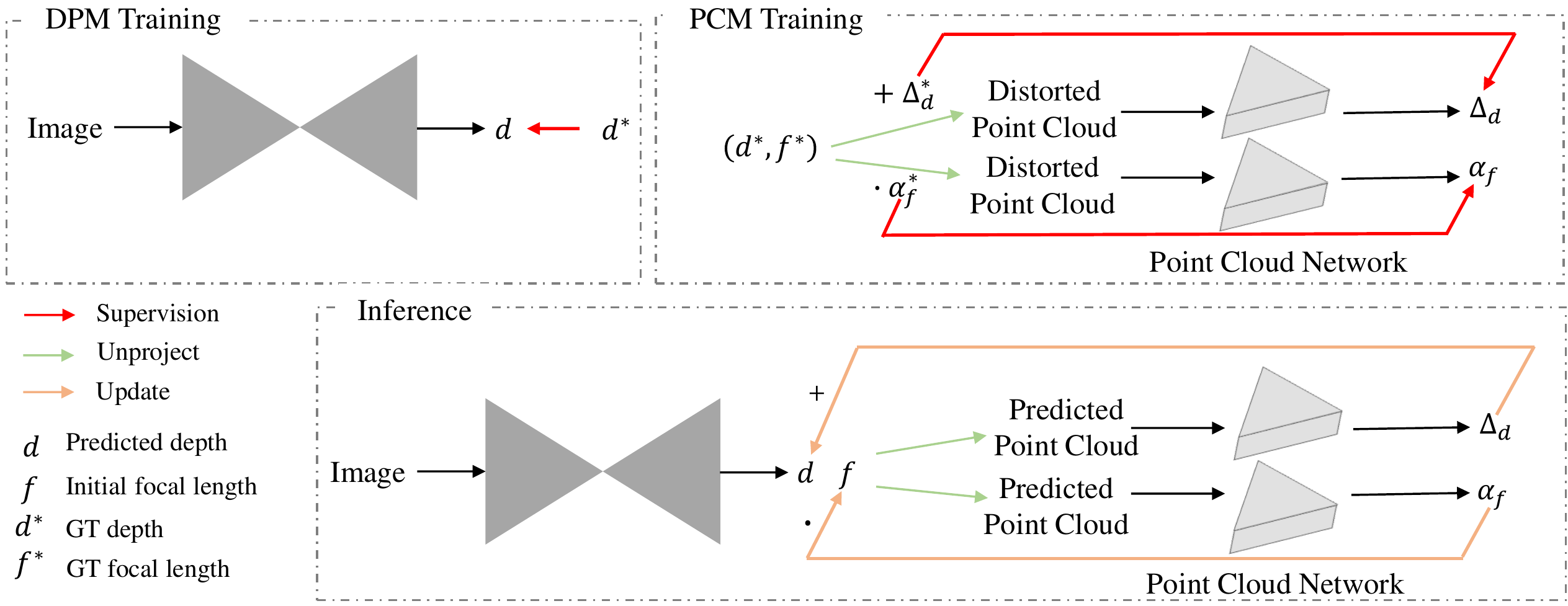}
\caption{\textbf{The overall pipeline of our method.}
During training, the depth prediction model (top left) and point cloud module (top right) are trained separately on different sources of data. During inference (bottom), the two networks are combined to predict depth $d$;  and %
the depth shift $\Delta_{d}$, %
the focal length $f\cdot\alpha_{f}$ using the predicted $ d $, %
which 
together %
enable 
an accurate scene shape reconstruction. Note that we employ point cloud networks to predict shift and focal length scaling factors separately. 
}
\label{Fig: pipeline}
\end{figure*}

\noindent\textbf{Camera intrinsic parameter estimation.} 
Recovering a camera's focal length is an important %
sub-task in
3D scene understanding. 
Traditional methods %
exploit 
reference objects such as a planar calibration grid \cite{zhang2000flexible} or vanishing points  \cite{deutscher2002automatic}, which can then be used to estimate %
the 
focal length. 
Other methods  \cite{hold2018perceptual, workman2015deepfocal} propose a data-driven approach where a CNN recovers the focal length 
directly from an image.
Here, 
our point cloud module
estimates the focal length directly in 3D, which %
is an easier task than operating on %
the images directly.

\noindent\textbf{Depth completion.} 
Depth completion aims to %
predict 
a dense depth map from a sparse or an incomplete depth. According to the type of the sparse depth, methods  \cite{cheng2020cspn++, cheng2019learning, park2020non, senushkin2020decoder, zhang2018deepdepth, WeiTRO2022} can be categorized into two main classes. 
The %
LiDAR for automotive vehicles or time-of-flight (ToF) for %
smartphones can only provide sparse depths with only a few hundred of pixels. Several methods  \cite{cheng2020cspn++, cheng2019learning, park2019deepsdf} %
leverage such sparse information. %
Park~\etal  \cite{park2020non} propose a non-local spatial propagation network to aggregate relevant information based on the predicted spatially-varying affinities,
demonstrating 
state-of-the-art performance on NYU and KITTI. 
By contrast, commodity-level RGBD cameras, such as Kinect and RealSense, can produce a more complete depth map but still miss data in some regions which are too glossy, bright, thin, close, or far from the camera. %
Some methods  \cite{herrera2013depth, zuo2016explicit} %
apply 
inpainting methods to fill the depth holes. Zhang~\etal  \cite{zhang2018deepdepth} use the surface normals and occlusion boundaries for depth completion and optimize global surface structures from those predictions with soft constraints provided by observed depths.

\section{Our Methods}
Our two-stage pipeline for  3D shape estimation from single images  is %
shown 
in Fig.~\ref{Fig: pipeline}. 
It %
consists 
of a depth recovery module and a point cloud module. 
The two modules are trained separately on different data sources and are then combined at the inference time. 
When there is only a single image input, the depth recovery module outputs a depth map  \cite{yin2020diversedepth} with unknown scale and shift in relation to the true metric depth map. In contrast, if some sparse depth points are available, the depth recovery module will take both RGB image and sparse points as input, and outputs the metric depth. 
The point cloud module takes as input a distorted 3D point cloud that is computed using a predicted depth map $d$ and an initial estimation of the focal length $f$,\footnote{This initial value does not need to be very accurate.} and outputs shift adjustments to the depth map and the focal length to improve the geometry of the reconstructed 3D scene shape. 
We describe the %
two modules next. 

\subsection{Point Cloud Module}
\label{sec:3D_shape}

\begin{figure}[t]
\centering
\includegraphics[width=.93999\linewidth]{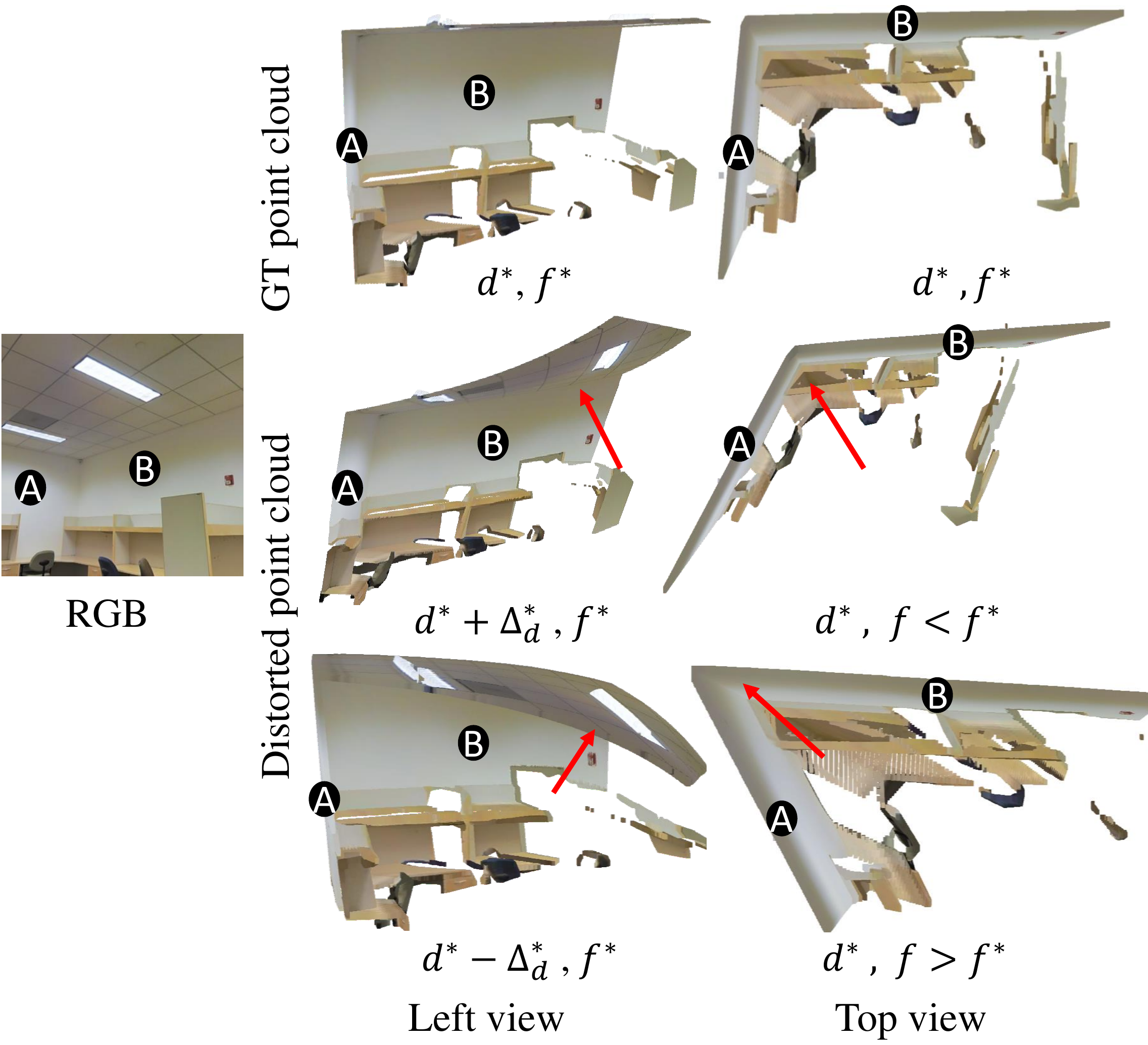}
\caption{\textbf{Illustration of the distorted 3D shape caused by incorrect shift and focal length}. A ground-truth depth map is projected in 3D, which can create the ground truth point cloud (see the first row). {\tt A} and {\tt B} annotate the walls. When the focal length is incorrectly estimated ($f>f^{*}$ or $f<f^{*}$), we observe significant structural distortion, \textit{e.g.},  see the angle between two walls {\tt A} and {\tt B} (see the third column). Second column: a shift ($d^{*}+\Delta_{d}$ or $d^{*}-\Delta_{d}$) also causes the shape distortion, see the roof. Note that different distortions are caused by the negative or positive shift. }
\label{Fig: focal length and shift affect shape.}
\vspace{-1.5em}
\end{figure}

We assume a pinhole camera model for the 3D point cloud reconstruction, which means that the un-projection from 2D coordinates and depth to 3D points is:
\begin{equation}
    \left\{\begin{matrix}x =&\hspace{-0.3cm} \frac{u-u_{0}}{f}d
    \\ y =&\hspace{-0.3cm} \frac{v-v_{0}}{f}d
    \\ z =&\hspace{-0.3cm} \hspace{0.75cm}d
    \end{matrix}\right.
\label{eq: 3D point cloud reconstruction}
\end{equation}
where $(u_{0}, v_{0})$ are the camera optical center;
$f$ is the focal length, and $d$ is the depth. 
The focal length affects the point cloud shape as it scales $x$ and $y$ coordinates, but not $z$.
Similarly, a depth shift %
$\Delta_{d}$ %
affects  the $x$, $y$, and $z$ coordinates non-uniformly, which %
results in shape distortions. 

For a human observer, these distortions are immediately recognizable when viewing the point cloud at an oblique angle, although they cannot be observed by looking at a depth map alone. In Fig.~\ref{Fig: focal length and shift affect shape.}, we can see that a shift for the depth will cause planes camber, while the focal length will change the angle between two planes (see the last column). 

As a result, we propose to \textit{directly analyze the point cloud to %
estimate 
the unknown shift and focal length, instead of working with 2D images}.
We tried several network architectures that take unstructured 3D point clouds as input and found that the recent PVCNN  \cite{liu2019pvcnn} performs
well for this task. Thus,  we build our method on the PVCNN 
architecture. %
During training, a perturbed input point cloud with incorrect shift and focal length is synthesized by perturbing the known ground-truth depth shift and focal length. 
The ground-truth depth $d^{*}$ is transformed by a shift $\Delta^{*}_{d}$ drawn from $\mathcal{U}(-0.25, 0.8)$, and the ground truth focal length $f^*$ is transformed by a scale $\alpha^{*}_{f}$ drawn from $\mathcal{U}(0.6, 1.25)$ to keep the focal length positive and non-zero.

When recovering the depth shift, the perturbed 3D point cloud %
$\mathcal{F}(u_{0}, v_{0}, f^{*}, d^{*} + \Delta^{*}_{d})$ is given as input to the shift point cloud network $\mathcal{N}_{d}(\cdot)$, trained with the objective: 
\begin{equation}
    L = \min_{\theta} 
    \left  | \mathcal{N}_{d}(\mathcal{F}(u_{0}, v_{0}, f^{*}, d^{*} 
           + 
        \Delta^{*}_{d}), \theta) - \Delta^{*}_{d} 
    \right | 
\label{eq: regress shift}
\end{equation}
where $\theta$ are network weights and $f^{*}$ is the true focal length, $\mathcal{F}(\cdot)$ is the mapping defined in Eq.~\eqref{eq: 3D point cloud reconstruction}. 

Similarly, when recovering the focal length, the point cloud $\mathcal{F}(u_{0}, v_{0}, \alpha^*_f f^*, d^{*})$ is fed to the focal length point cloud network $\mathcal{N}_{f}(\cdot)$, trained with the objective:
\begin{equation}
    L = \min_{\theta}\left | \mathcal{N}_{f}(\mathcal{F}(u_{0}, v_{0}, \alpha^{*}_{f}f^{*}, d^{*}), \theta) - \alpha^{*}_{f} \right | 
\end{equation}

During the inference, the ground-truth depth is replaced with the predicted affine-invariant depth $d$, which is normalized to $[0, 1]$ prior to the 3D reconstruction. 
We use an initial guess of focal length $f$, giving us the reconstructed point cloud $\mathcal{F}(u_{0}, v_{0}, f, d)$, which is fed to $\mathcal{N}_{d}(\cdot)$ and $\mathcal{N}_{f}(\cdot)$ to predict the shift $\Delta_{d}$ and focal length scaling factor $\alpha_{f}$ respectively. 
In our experiments, we simply use an initial focal length with a field of view (FOV) of $60^{\circ}$. 
We have also tried to employ a single network to predict both the shift and the scaling factor, but have empirically found that two separate networks 
achieve better performance.

\subsection{Monocular Depth Prediction Module}

When there is only an RGB image input, we propose a monocular depth prediction module (DPM, see Fig.~\ref{Fig: pipeline}), which takes $I_{rgb}$ as input 
and produces an affine-invariant depth map $d$. We train our depth prediction on multiple data sources including high-quality LiDAR sensor data, %
medium-quality 
calibrated stereo data, and low-quality web stereo data\cite{Ranftl2020, wang2019web,  xian2020structure} (see Sec.~\ref{sec:data}).
As these datasets have varied depth ranges and web stereo datasets contain  unknown depth scale and shift, we propose an image-level normalized regression (ILNR) loss to %
facilitate 
the training. 
Moreover, we propose a pair-wise normal regression (PWN) loss to %
exploit 
local geometry information. 

\noindent\textbf{Image-level normalized regression loss.} 
Depth maps of different data sources can have varied depth ranges. Normalization is a critical step to transform data with variable ranges to a comparable range where large features no longer dominate smaller features  \cite{singh2019investigating}.  Therefore, we propose to normalize the data to make the model training easier. Simple Min-Max normalization  \cite{ garcia2015data, singh2019investigating} is sensitive to depth value outliers. For example, a large value at a single pixel will affect the rest of the depth map after the Min-Max normalization. We investigate more robust normalization methods and propose a simple but effective image-level normalized regression loss for mixed-data training.

Our image-level normalized regression loss transforms each ground-truth depth map to a similar numerical range based on its individual statistics. 
To reduce the effect of outliers and long-tail residuals, we combine {$\rm tanh$}   normalization  \cite{singh2019investigating} with a trimmed Z-score normalization, after which we can simply apply a pixel-wise mean average error (MAE) between the prediction and the normalized ground-truth depth maps.
The ILNR loss is formally defined as follows.
\begin{align*}
\label{normalization loss}
  L_\text{ILNR} &= \frac{1}{N}{\sum_{i}^{N}\left|d_{i} - \overline{d}^{*}_{i}\right| + \left|\text{tanh}(\nicefrac{d_{i}}{100}) - \text{tanh}(\nicefrac{\overline{d}^{*}_{i}}{100}) \right|}
\end{align*}
where $\overline{d}^{*}_{i} =  \nicefrac{ ( d^{*}_{i} - \mu_\text{trim} )} {\sigma_\text{trim} }$ and 
$\mu_{\rm trim}$ and $\sigma_{\rm trim}$ are the mean and the standard deviation of a trimmed depth map which has the nearest and farthest $10\%$ of pixels removed. $d$ is the predicted depth, and $d^{*}$ is the ground-truth depth map.

We have tested a number of other normalization methods such as Min-Max normalization  \cite{singh2019investigating}, Z-score normalization  \cite{fukunaga2013introduction}, and median absolute deviation normalization \cite{singh2019investigating}. In our experiments, we %
observe 
that our proposed ILNR loss achieves the best performance and generalization.

\noindent\textbf{Pair-wise normal loss.} 
Surface normals are an important geometric property, which has been shown to be a complementary modality to depth  \cite{silberman2012indoor}.
Many methods have been proposed to use normal constraints to improve the depth quality, such as the virtual normal loss  \cite{Yin2019enforcing}. 
However, as the virtual normal only leverages global structure, it %
may not help improve the local geometric quality, such as edges and planes. Recently, Xian \etal  \cite{xian2020structure} proposed a structure-guided ranking loss, which can improve edge sharpness. Inspired by these methods, we follow their sampling method but enforce the supervision in the surface normal space. Moreover,  our samples include not only edges but also planes. Our proposed pair-wise normal (PWN) loss can better constrain both the global and local geometric relations.

The detailed sampling method is described here. The first step is to locate image edges. Following \cite{Yin2019enforcing}, we calculate the surface normal from the depth map with the local least squares fitting method. The Sobel edge detector is applied to %
find 
edges from the surface normal map and the input image. At each edge point, we then sample pairs of points on both sides of the edge. The ground-truth normals for these points are $\mathscr{N}^{*} = \{(\bm{n}^{*}_{A}, \bm{n}^{*}_{B})_{i} | i=0,...,n\}$, while the predicted normals are $\mathscr{N} = \{(\bm{n}_{A}, \bm{n}_{B})_{i} | i=0,...,n\}$. Before calculating the predicted surface normal, we align the predicted depth and the ground-truth depth with a scale and shift factor, which are retrieved by the least squares fitting  \cite{Ranftl2020}. To locate the object boundaries and %
intersection edge of planes, where the normals change significantly, we set the angle difference of two normals greater than ${\rm arccos}(0.3)$. To balance the samples, we also get some negative samples, where the angle difference is smaller than  ${\rm arccos}(0.95)$ and they are also detected as edges on the input image. The sampling method is %
as follows: 
\begin{equation}
    \mathscr{S}_{1} = \{\bm{n}^{*}_{A} \cdot \bm{n}^{*}_{B} > 0.95, \bm{n}^{*}_{A} \cdot \bm{n}^{*}_{B} < 0.3  | {(\bm{n}^{*}_{A}, \bm{n}^{*}_{B})_{i}} \in \mathscr{N}^{*} \}
\label{eq: constraints}
\end{equation}

Apart from samples on edges, we also sample points on the same plane to improve the quality of predicted planes. We enforce the co-plane supervision on samples that are on the same plane. Specifically, we calculate their normals and enforce their normals to be the same. 
We tried the surface normal and virtual normal~\cite{yin2021virtual} on samples for the co-plane constraint and found that they have a similar performance. During training, we sample around $~100$K paired points per training sample on average. The samples are %
$\{(A_{i}, B_{i}), i=0,...,N\}$.
The PWN loss is:
\begin{equation}
\label{normalization loss}
\begin{aligned}
  L_\text{PWN} &= \frac{1}{N}\sum_{i}^{N}\left|n_{Ai} \cdot n_{B_i} - n^{*}_{Ai}  \cdot  n^{*}_{Bi} \right|
\end{aligned}
\end{equation}
where $n^{*}$ denotes ground truth surface normals. Note that if points are on the same plane, $n^{*}_{Ai}  \cdot  n^{*}_{Bi} =1$.
As this loss accounts for both local and global geometry, we find that it improves the overall reconstructed shape.

\begin{figure}[t]   
\centering
\includegraphics[width=\linewidth]{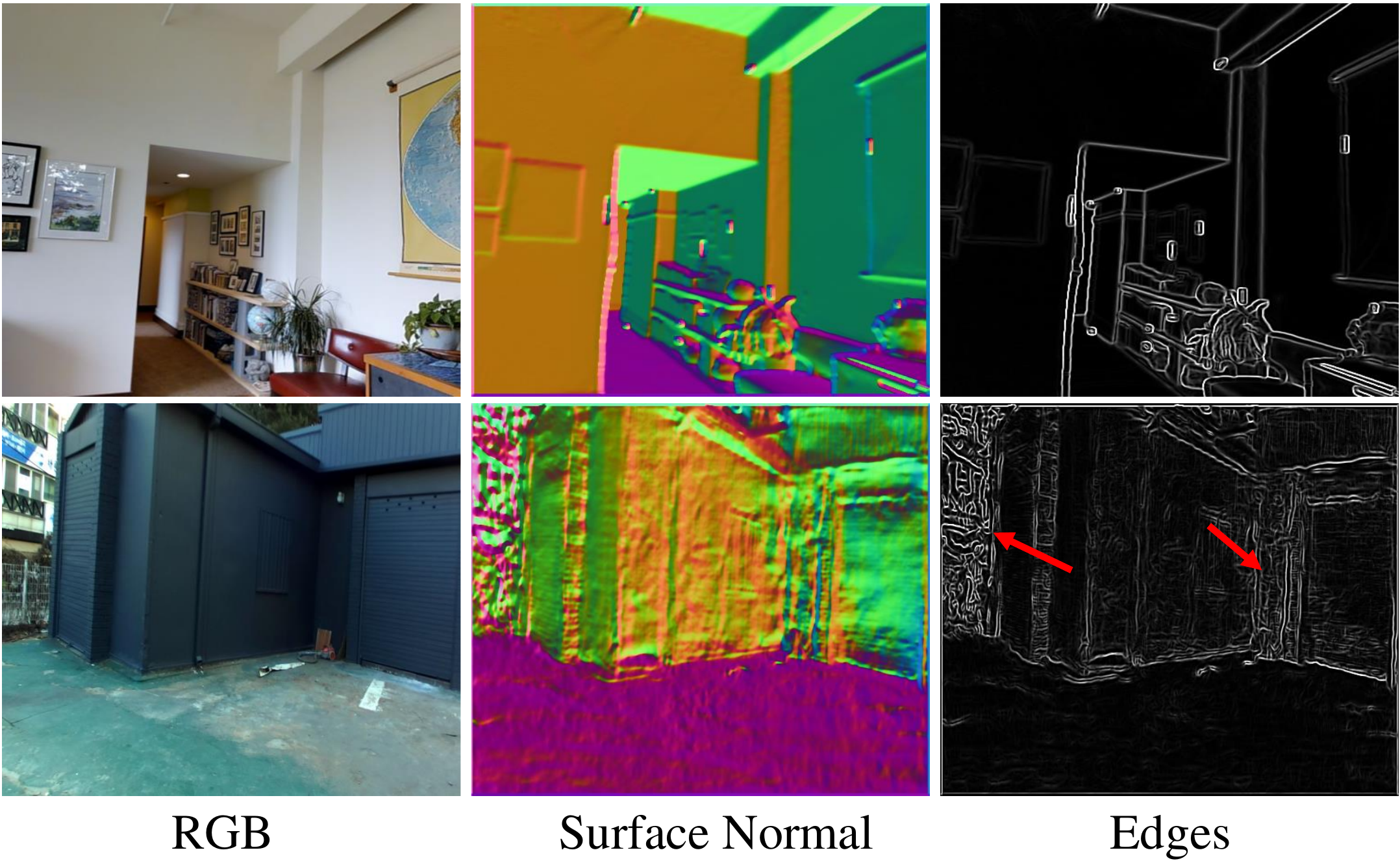}
\caption{Detected edges on high-quality (Taskonomy~\cite{zamir2018taskonomy}, the first row) and medium-quality data (DIML~\cite{kim2018deep}, the second row). There are more artifacts on DIML (see red arrows). }
\label{Fig: cmp of edges.}
\end{figure}

\noindent\textbf{Creating instance planes.} To enforce the PWN loss on planes, we create instance planes on the high-quality LiDAR data (Taskonomy \cite{zamir2018taskonomy}) and medium-quality data (DIML~\cite{kim2018deep}). On Taskonomy, we use the least-squares  fitting method~\cite{Yin2019enforcing} to %
compute 
surface normals from depths. Then we employ the DBSCAN algorithm~\cite{DBCAN} to cluster planes that have the same surface normals. As the dataset DIML is much noisier than Taskonomy, we cannot use such fitting methods to obtain instance planes. Fig.~\ref{Fig: cmp of edges.} shows the comparison. We use  \cite{deeplabv3plus2018} to segment roads, which we assume to be planar regions.

Finally, we also use a multi-scale gradient loss (MSG)  \cite{li2018megadepth} and structure-guided ranking loss (SR)  \cite{xian2020structure}.
The MSG loss is as follows:
\begin{equation}
L_\text{MSG}=\frac{1}{N}\sum_{k=1}^{K}\sum_{i=1}^{N}\left |\bigtriangledown ^{k}_{x}d_{i} -\bigtriangledown ^{k}_{x}\overline{d}^{*}_{i}    \right | + \left |\bigtriangledown ^{k}_{y}d_{i} -\bigtriangledown ^{k}_{y}\overline{d}^{*}_{i}    \right |
\label{multi-scale gradient loss}
\end{equation}

The structure-guided ranking loss is as follows:
\begin{equation}
    L_\text{SR} = \frac{1}{N}\sum_{i=0}^{N} \left\{\begin{matrix} \log 
    \left (1 + \exp
        \left [ -l\left (d_{i0} - d_{i1}  \right ) \right ]  \right ); \;\; l\neq 0
\\ \left ( d_{i0} - d_{i1} \right )^{2}, \;\; l = 0, 
\end{matrix}\right.
\end{equation}
where $
l=\left\{\begin{matrix}+1, \;\;\; {\rm if} \;  d^{*}_{i0} / d^{*}_{i1} 
\geq 
1+\tau; 
\\ -1,  \;\;\; {\rm if} \;
        d^{*}_{i1} / d^{*}_{i0} 
\geq 
1 + \tau; 
\\   \!\!\!    0, \;\;\;   \rm  otherwise.
\end{matrix}\right.$. 

\noindent\textbf{%
The training strategy for mixed datasets.}
We mix $5$ %
datasets to train the depth model. Based on their depth quality, they are categorized to high-quality data (Taskonomy  \cite{zamir2018taskonomy} and 3d ken burns  \cite{Niklaus_TOG_2019}), medium-quality data (DIML  \cite{kim2018deep}), and low-quality data (Holopix50K  \cite{hua2020holopix50k} and HRWSI  \cite{xian2020structure}).  Some examples of such datasets are illustrated in Fig.~\ref{Fig: cmp of data.}.

For the low-quality web-stereo data, as their inverse depths $d_{inv}$ have unknown scale and shift, \ie, $d_{inv} = s  \cdot d_{inv}^{*} + \Delta_{d\_inv}$, the depth map ($d = 1/d_{inv} = 1/(s   \cdot d_{inv}^{*} + \Delta_{d\_inv})$) can only demonstrate the relative depth relations. Therefore, we only enforce the structured-guided ranking loss on 
those data in the depth space. In Fig.~\ref{Fig: cmp of data.}, we can see the reconstructed point cloud of low-quality data has noticeable distortions.

For the medium-quality data, such as DIML  \cite{kim2018deep}, we enforce the proposed image-level normalized regression loss, multi-scale gradient loss, and ranking loss. As depths contain much noise in local regions (see Fig.~\ref{Fig: cmp of edges.}), enforcing the pair-wise normal regression loss on noisy edges will cause many artifacts. Therefore, we enforce the pair-wise normal regression loss only on planar regions.

\iffalse
\begin{figure}[t]   
\centering
\includegraphics[width=\linewidth]{./Figs/data_cmp.pdf}
\caption{\textcolor{red}{Some examples of training data. High-quality and medium-quality data are from Taskonomy and DIML respectively. Low-quality data is from HRWSI, of which the unprojected point cloud has significant distortion.} }
%
\label{Fig: cmp of data.}
\end{figure}
\fi

For the high-quality data, we %
apply 
all the three loss terms. 
The overall loss functions for different datasets are %
reported 
in  Table~\ref{Tab: losses for different datasets}.

\begin{table}[]
\resizebox{1\linewidth}{!}{
\begin{tabular}{l|lllll}
\toprule[1pt]
                    & $L_\text{SR}$ & $L_\text{ILNR}$ & \begin{tabular}[c]{@{}l@{}}$L_{\rm PWN}$\\ (Edges)\end{tabular} & \begin{tabular}[c]{@{}l@{}}$L_{\rm PWN}$\\ (Planes)\end{tabular}  & $L_\text{MSG}$  \\ \hline
High-quality (Taskonomy~\cite{zamir2018taskonomy}, 3D Ken Burn~\cite{Niklaus_TOG_2019})   &\checkmark &\checkmark &\checkmark &\checkmark &\checkmark \\
Medium-quality (DIML~\cite{kim2018deep}) &\checkmark &\checkmark & &\checkmark &\checkmark \\
Low-quality (HRWSI~\cite{xian2020structure}, Holopix~\cite{hua2020holopix50k})    &\checkmark   &   &    &    & \\ \toprule[1pt]
\end{tabular}}
\caption{Losses 
used 
for different datasets based on their depth quality.\label{Tab: losses for different datasets}}
\vspace{-1 em}
\end{table}

\subsection{Depth Completion Module}
To recover accurate metric reconstruction, we propose to combine a depth completion module with PCM. Note that we only need the PCM to predict the focal length.
Existing depth completion methods are mainly classified into two categories according to the input sparsity pattern: depth inpainting methods that fill large holes~\cite{zhang2018deepdepth, senushkin2020decoder, huang2019indoor}, and sparse depth completion methods that fill depths with only hundreds of sparse points~\cite{cheng2020cspn++, park2020non, xu2019depth, qiu2019deeplidar, cheng2019learning}. 
Current methods~\cite{park2020non, cheng2020cspn++, cheng2018depth, qiu2019deeplidar} can achieve impressive performance on a specific sparsity pattern and scene, \textit{e.g.}, on NYU~\cite{silberman2012indoor}, KITTI~\cite{Uhrig2017THREEDV}, and Matterport3D~\cite{Matterport3D}. However, in real-world scenarios, the problem is that the specific sparsity pattern may be unknown,
as it is conditioned on hardware, software, and the configuration of the scene itself. Due to the above reasons, multiple models have to be trained to solve various sparse depth situations. Furthermore, outliers and
depth sensor noises are unavoidable in any depth acquisition
method. Most of previous methods only take an RGB image
and a sparse depth as the input, and they do not have any extra source of information with which it could distinguish the
outliers. 

Our proposed mixed-data training strategy is an effective %
approach 
to improve the generalization of the DPM. During training, we synthetically create the sparse depth input by sampling from the ground-truth depth. We create a few different  sparsity patterns, including 1) Uniform sampling. We sample  uniformly  distributed  points, from hundreds to thousands of points; 2) Sampling feature points. we use a feature detector to sample points from textured regions and edge or corner parts. In our experiment, FAST~\cite{rosten2006machine} feature detector is employed; 3) 
Creating holes. We mask the depth by using a random polygonal region or at a certain distance. Examples of sparsity patterns are shown in Figure~\ref{Fig: sdepth patterns}. Furthermore, to improve the robustness to noisy inputs, we propose to input a data prior, which is from our DPM, to help resolve incorrect constraints when there is a significant discrepancy between the two. In order to encourage the network to learn this, we add outliers during training. Specifically, we randomly sample several points and scale their depth by a random factor from 0.1 to 2.

During training, we only use the high-quality data (Taskonomy~\cite{zamir2018taskonomy}) and medium-quality data (DIML~\cite{kim2018deep}) to train the model. The RGB image, sparse depth, and guidance map are concatenated to be input to ESANet-R34-NBt1D~\cite{seichter2021efficient}  network. The framework is shown in Figure~\ref{Fig: depth completion framework. }. Heterogeneous losses are enforced on two datasets, see Table~\ref{Tab: losses for depth completion} for details. $L_{\text {MAE} }$ is the mean average error loss.

\begin{table}[]
\resizebox{1\linewidth}{!}{
\begin{tabular}{l|lllll}
\toprule[1pt]
                    & $L_\text{SR}$ & $L_\text{MAE}$ & \begin{tabular}[c]{@{}l@{}}$L_{\rm PWN}$\\ (Edges)\end{tabular} & \begin{tabular}[c]{@{}l@{}}$L_{\rm PWN}$\\ (Planes)\end{tabular}  \\ \hline
High-quality (Taskonomy~\cite{zamir2018taskonomy})   &\checkmark &\checkmark &\checkmark &\checkmark\\
Medium-quality (DIML~\cite{kim2018deep}) &\checkmark &\checkmark & &\checkmark \\ \toprule[1pt]
\end{tabular}}
\caption{Losses 
used 
on different datasets for the depth completion.\label{Tab: losses for depth completion}}
\vspace{-2 em}
\end{table}

\section{Experiments}

\noindent\textbf{Datasets and implementation details.}
\label{sec:data}
To train the PCM, we sample $100$K Kinect-captured depth maps from ScanNet, $114$K LiDAR-captured depth maps from Taskonomy, and $51$K synthetic depth maps from the 3D Ken Burns paper  \cite{Niklaus_TOG_2019}.
We train the network using SGD with a batch size of $40$, an initial learning rate of $0.24$, and a learning rate decay of $0.1$. For parameters specific to PVCNN, such as the voxel size, we follow the original work  \cite{liu2019pvcnn}.

To train the DPM, we sample  $114$K RGBD pairs from LiDAR-captured Taskonomy  \cite{zamir2018taskonomy}, $51$K synthetic RGBD pairs from the 3D Ken Burns paper  \cite{Niklaus_TOG_2019},  $121$K RGBD pairs from calibrated stereo DIML  \cite{kim2018deep}, $48$K RGBD pairs from web-stereo Holopix50K  \cite{hua2020holopix50k}, and $20$K web-stereo HRWSI  \cite{xian2020structure} RGBD pairs.
Note that %
for 
the ablation study about the effectiveness of PWN and ILNR, we sample  a smaller dataset which is composed of $12$K images from Taskonomy, $12$K images from DIML, and $12$K images from HRWSI. 
During training, $1000$ images are withheld from all datasets as a validation set.
We use the depth prediction architecture proposed in Xian~\etal \cite{xian2020structure}, which consists of a standard backbone for feature extraction (\textit{e.g.}, ResNet50  \cite{he2016deep} or ResNeXt101  \cite{xie2017aggregated}), followed by a decoder, and train it using SGD with a batch size of $40$, an initial learning rate $0.02$ for all layer, and a learning rate decay of $0.1$. 
Images are resized to $448$$\times$$448$, and flipped horizontally with a $50\%$ chance.
Following \cite{yin2020diversedepth}, we load data from different datasets evenly for each batch. More details about training refer to the appendix.

\noindent\textbf{Evaluation details.}
The accuracy of focal length prediction  is evaluated on 2D-3D-S  \cite{armeni2017joint} following  \cite{hold2018perceptual}. 
Furthermore, to evaluate the accuracy of the reconstructed 3D shape, we use the Locally Scale-Invariant RMSE (LSIV)  \cite{chen2020oasis} metric on both OASIS  \cite{chen2020oasis} and 2D-3D-S  \cite{armeni2017joint}. It is consistent with the previous work  \cite{chen2020oasis}. The OASIS  \cite{chen2020oasis} dataset only has the ground-truth depth on some small regions, while 2D-3D-S has the ground truth for the whole scene.

\begin{table}[t]
\centering
\resizebox{1\linewidth}{!}{%
\begin{tabular}{l|ccccc}
\toprule[1pt]
\multicolumn{1}{c|}{\multirow{2}{*}{Method}} & ETH3D & NYU & KITTI & Sintel & DIODE \\
\multicolumn{1}{c|}{}                        & \multicolumn{5}{c}{AbsRel$\downarrow$}           \\ \hline
Baseline (Aligned scale)    & $23.7$    &$25.8$     &$23.3$   &$47.4$   &$46.8$    \\
Recovered shift + aligned scale & $\textbf{15.9}$    &${15.1}$    &${17.5}$  &${40.3}$   &${36.9}$  \\
Aligned scale \& shift & ${17.1}$    &$\textbf{9.1}$    &$\textbf{14.3}$  &$\textbf{31.9}$   &$\textbf{27.1}$ \\ \toprule[1pt]
\end{tabular}}
\caption{Effectiveness of recovering the shift from 3D point clouds with the PCM. Compared with the baseline, the AbsRel$\downarrow$ is much lower after recovering the depth shift over all test sets.\label{Tab: effectiveness of shift prediction. }}
\vspace{-1em}
\end{table}

\begin{figure}[t]   
\centering
\includegraphics[width=\linewidth]{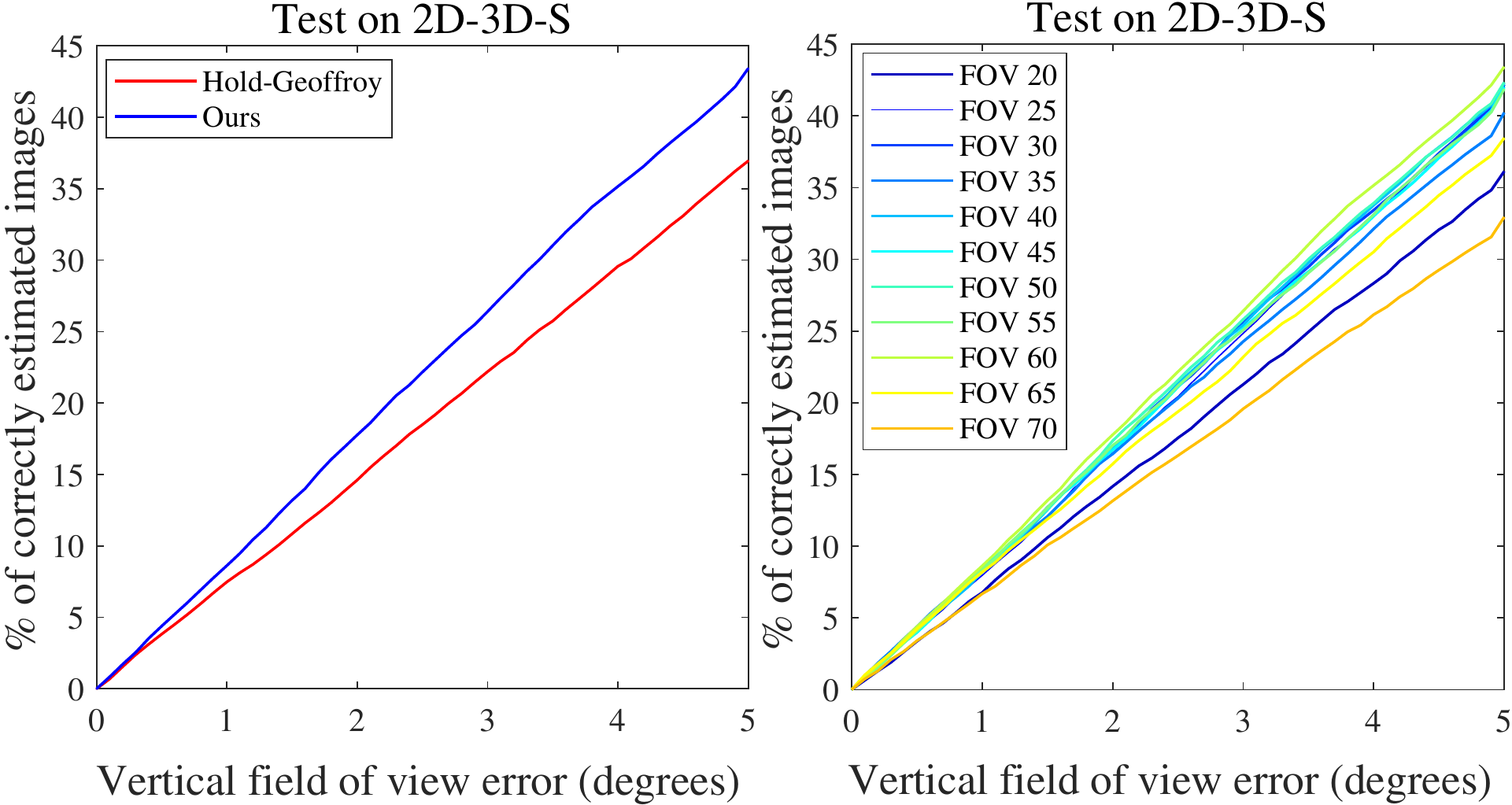}
\caption{Zero-shot evaluation of the focal length on the 2D-3D-S dataset. 
Following~\cite{hold2018perceptual}, we use the percentage of correctly estimated images for evaluation.
Left: our method outperforms Hold-Geoffroy~\etal  \cite{hold2018perceptual}. Right: we experiment on the effect of the initialization of field of view (FOV). Our method remains robust across different initial FOVs, with a slight degradation in quality %
beyond 
$25^{\circ}$ and $65^{\circ}$.}
\vspace{-1em}
\label{Fig: cmp of focal length prediction.}
\end{figure}

\begin{figure*}[t]
\centering
\includegraphics[width=.9\linewidth]{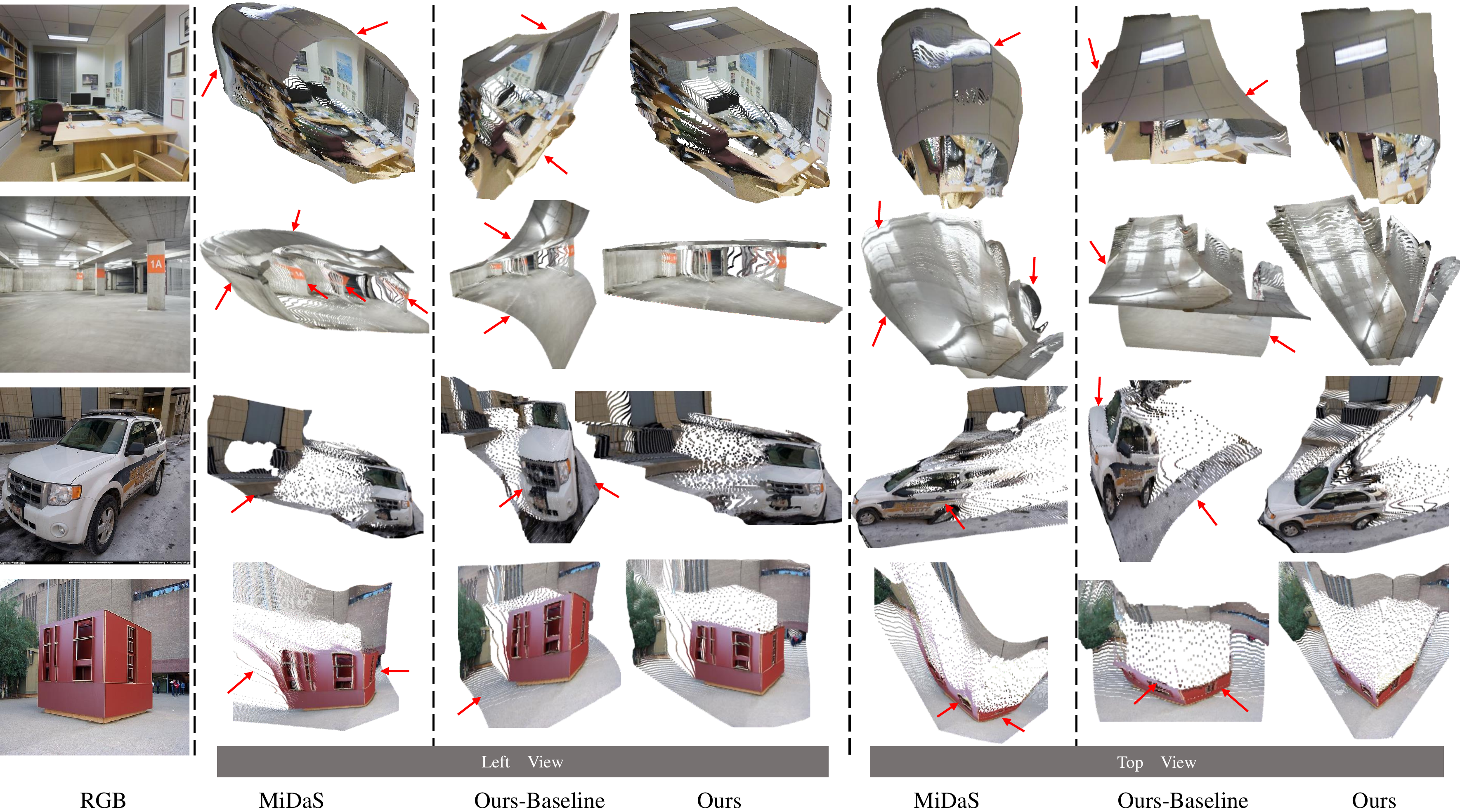}
\caption{
\textbf{Qualitative comparison.} 
We compare the reconstructed 3D shape of our method with several baselines. As MiDaS  \cite{Ranftl2020} does not estimate the focal length, we use the focal length recovered from \cite{hold2018perceptual} to convert the predicted depth to a point cloud.
``Ours-Baseline'' does not recover the depth shift or focal length and uses an orthographic camera, while ``Ours'' recovers the shift and focal length. Our method can better reconstruct the 3D shape, especially at edges and planar regions (see arrows).}
\label{Fig: cmp of SOTA 3D shape. }
\vspace{-1em}
\end{figure*}

To evaluate the robustness  of our depth prediction method, we %
test on 
$9$ datasets which are unseen during training, including YouTube3D  \cite{chen2019learning}, OASIS  \cite{chen2020oasis}, NYU  \cite{silberman2012indoor}, KITTI  \cite{geiger2012we}, ScanNet  \cite{dai2017scannet}, DIODE  \cite{vasiljevic2019diode}, ETH3D  \cite{schops2017multi}, Sintel  \cite{Butler:ECCV:2012}, and iBims-1  \cite{Koch18:ECS}. 
On OASIS and YouTube3D, we use the Weighted Human Disagreement Rate (WHDR)  \cite{xian2018monocular} for evaluation. 
On other datasets, except for iBims-1, we use the absolute mean relative error (AbsRel$\downarrow$) and the percentage of pixels with $\delta_{1}=\text{max}(\frac{d_{i}}{d^{*}_{i}}, \frac{d^{*}_{i}}{d_{i}})<1.25$. 
We follow Ranftl~\etal  \cite{Ranftl2020} and align the scale and shift before evaluation. 

To evaluate the geometric quality of the depth, \ie, the quality of edges and planes, we follow  \cite{Niklaus_TOG_2019, xian2020structure} and evaluate the depth boundary error  \cite{Koch18:ECS} ($\varepsilon^\text{acc}_\text{DBE}, \varepsilon^\text{comp}_\text{DBE}$) as well as the planarity error  \cite{Koch18:ECS} ($\varepsilon^\text{plan}_\text{PE}, \varepsilon^\text{orie}_\text{PE}$) on iBims-1. 
$\varepsilon^\text{plan}_\text{PE}$ and $\varepsilon^\text{orie}_\text{PE}$ evaluate the flatness and orientation
of reconstructed 3D planes compared to the ground truth 3D planes respectively, while $\varepsilon^\text{acc}_\text{DBE}$ and $\varepsilon^\text{comp}_\text{DBE}$ demonstrate the localization accuracy and the sharpness of edges respectively. More details as well as a comparison of 
these test datasets are shown in  Table~\ref{Tab: testing data details}.

\subsection{3D Shape Reconstruction}

\noindent\textbf{Shift recovery.}
To evaluate the effectiveness of our depth shift recovery, we perform the zero-shot evaluation on $5$ datasets unseen during training.
We recover a 3D point cloud by unprojecting the predicted depth map and then compute the depth shift using our PCM. 
We then align the unknown scale  \cite{bian2019unsupervised, monodepth2} of the original depth and our shifted depth to the ground-truth and evaluate both using the AbsRel$\downarrow$ error.
The results are shown in  Table~\ref{Tab: effectiveness of shift prediction. }, where we see that, on all test sets, the AbsRel$\downarrow$ error is lower after recovering the shift. 
We also trained a standard 2D CNN to predict the shift given an image composed of the un-projected point coordinates, but this approach did not generalize well to samples from unseen datasets. 

\noindent\textbf{Focal length recovery.}
To evaluate the accuracy of our recovered focal length, we follow Hold-Geoffroy~\etal  \cite{hold2018perceptual} and test on the 2D-3D-S dataset, which is unseen during training for both methods. 
The model of  \cite{hold2018perceptual} is trained on the in-the-wild SUN360  \cite{xiao2012recognizing} dataset. 
We pre-define a degree error threshold in classifying whether the recovered focal length for one image is correct or not. Results are shown in Fig.~\ref{Fig: cmp of focal length prediction.}. We can see that our method demonstrates better generalization.
Note that PVCNN is very lightweight and only has $5.5$M  parameters, but shows promising generalization capability. It %
indicates that there is a much smaller domain gap between datasets in the 3D point cloud space than that in the image space where appearance variation can be large.

Furthermore, we analyze the effect of different initial focal lengths during inference. 
We set the initial field of view (FOV) from $20^{\circ}$ to $70^{\circ}$ and evaluate the accuracy of the recovered focal length, Fig.~\ref{Fig: cmp of focal length prediction.} (right). 
The experimental results demonstrate that our method is not particularly sensitive to different initial focal lengths.

\begin{table}[t]
\centering
\resizebox{\linewidth}{!}{%
\begin{tabular}{l|ll}
\toprule
\multirow{2}{*}{Method} & OASIS & 2D-3D-S \\
                        & LSIV $\downarrow$ & LSIV$\downarrow$  \\ \hline \hline  
\multicolumn{3}{c}{Orthographic Camera Model} \\ \hline \hline               
MegaDepth  \cite{li2018megadepth}               &$0.64$       &$2.68$       \\
MiDaS  \cite{Ranftl2020}               &$0.63$       &$2.65$       \\
Ours-DPM              &$0.63$       &$2.65$      \\ \hline \hline
\multicolumn{3}{c}{Pinhole Camera Model} \\ \hline \hline 
MegaDepth  \cite{li2018megadepth} + Hold-Geoffroy  \cite{hold2018perceptual} &$1.69$  &$1.81$      \\
MiDaS  \cite{Ranftl2020} + Hold-Geoffroy  \cite{hold2018perceptual}&$1.60$ &$0.94$       \\
MiDaS  \cite{Ranftl2020} + Ours-PCM &$1.32$ &$0.94$       \\
Ours                   &$\textbf{0.52}$       &$\textbf{0.80}$       \\ 
\bottomrule
\end{tabular}}
\vspace{0.4em}
\caption{Quantitative evaluation of the reconstructed 3D shape quality on OASIS and 2D-3D-S. Our method can achieve better performance than previous methods. Compared with the orthographic projection, our method using the pinhole camera model can obtain better performance. DPM and PCM refer to our depth prediction module and point cloud module, respectively.
\label{Tab: shape evaluation on OASIS}}
 \vspace{-1em}
\end{table}

\begin{figure*}[t]
\centering
\includegraphics[width=\linewidth]{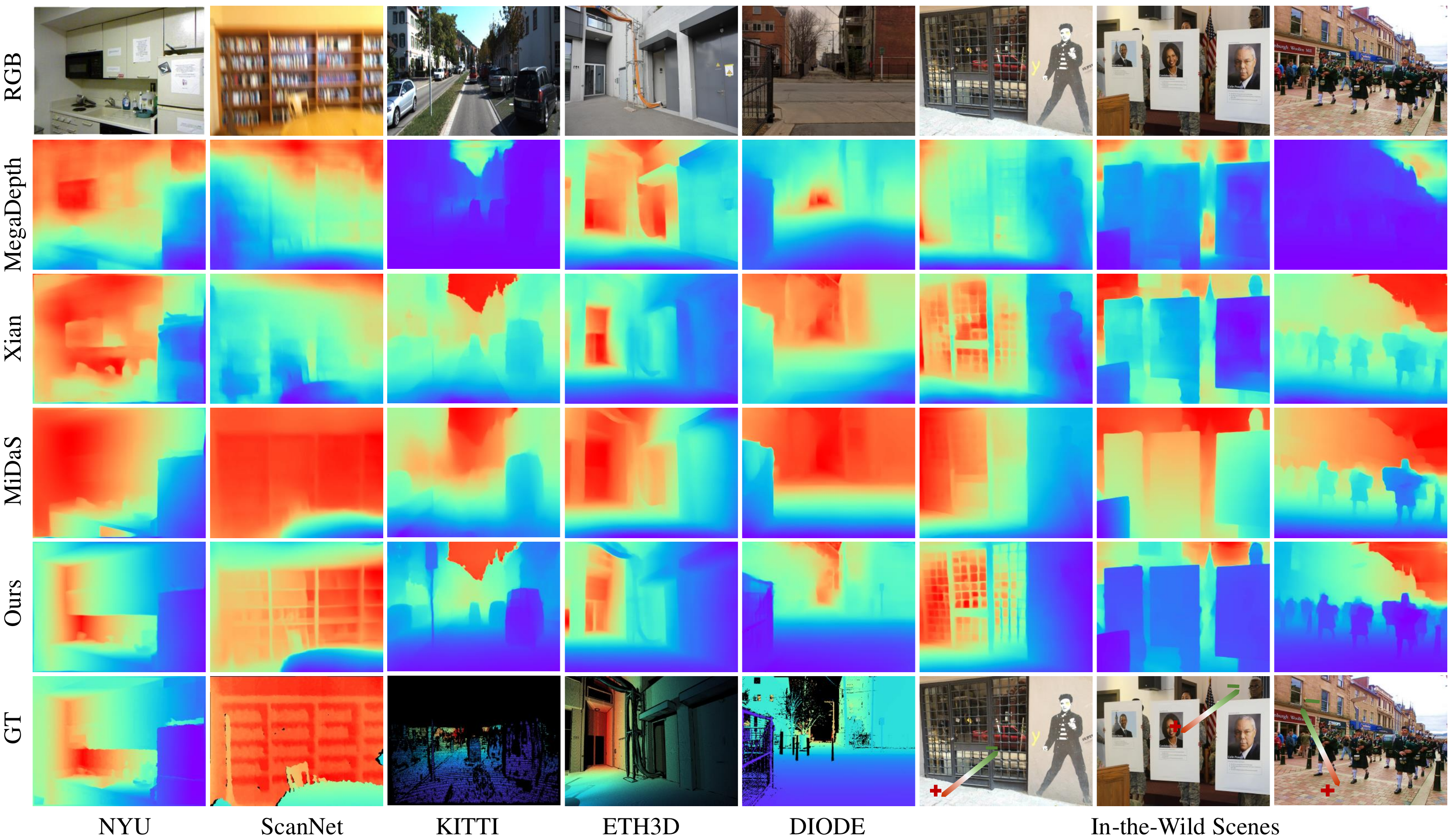}
\caption{
\textbf{Qualitative comparisons with state-of-the-art methods}, including MegaDepth  \cite{li2018megadepth}, Xian~\etal\cite{xian2020structure}, and MiDaS  \cite{Ranftl2020}. It shows that our method can predict more accurate depths at far locations and regions with complex details. In addition, we see that our method generalizes better to in-the-wild scenes.\label{Fig: cmp of SOTA. }}
\end{figure*}

\noindent\textbf{Evaluation of 3D shape quality.}
Following OASIS  \cite{chen2020oasis}, we use LSIV for the quantitative comparison of recovered 3D shapes on the OASIS  \cite{chen2020oasis} dataset and the 2D-3D-S  \cite{armeni2017joint} dataset. 
OASIS only provides the ground truth point cloud on small regions, while 2D-3D-S covers the whole 3D scene. %
Following OASIS  \cite{chen2020oasis}, we evaluate the reconstructed 3D shape with two different camera models, \ie, the orthographic projection camera model  \cite{chen2020oasis} (infinite focal length) and the (more realistic) pinhole camera model. 
As MiDaS  \cite{Ranftl2020} and MegaDepth  \cite{li2018megadepth} do not estimate the focal length, we use the focal length recovered from Hold-Geoffroy  \cite{hold2018perceptual} to convert the predicted depth to a point cloud.
We also evaluate a baseline using MiDaS instead of our DPM with the focal length predicted by our PCM (``MiDaS + Ours-PCM'').
Table~\ref{Tab: shape evaluation on OASIS} shows that, with an orthographic projection, our method (``Ours-DPM'') performs  roughly as well as existing state-of-the-art methods. However, for the pinhole camera model, our combined method significantly outperforms existing approaches. Furthermore, ``MiDaS + Ours-PCM'' and ``MiDaS + Hold-Geoffroy'' show that our PCM is able to generalize to different depth prediction methods. 

A qualitative comparison of the reconstructed 3D shape for in-the-wild scenes is shown in Fig.~\ref{Fig: cmp of SOTA 3D shape. }. 
It demonstrates that our model can recover significantly more accurate 3D scene shapes. For example, planar structures such as walls, floors, and roads are much flatter in our reconstructed scenes, and the angles between surfaces (\eg, walls) are also more realistic. Also, 
the shape of the car has fewer distortions.
\begin{table}[t]
\small
\centering
\begin{threeparttable}
\resizebox{1\linewidth}{!}{%
\begin{tabular}{l|ccccc}
\toprule[1pt]
\multirow{2}{*}{Method} & \multicolumn{4}{c}{iBims-1}                                    \\ \cline{2-6} 
        & $\varepsilon^\text{acc}_\text{DBE}\downarrow$ & \multicolumn{1}{l|}{$\varepsilon^\text{comp}_\text{DBE}\downarrow$} & $\varepsilon^\text{plan}_\text{PE}\downarrow$ & \multicolumn{1}{l|}{$\varepsilon^\text{orie}_\text{PE}\downarrow$} &AbsRel$\downarrow$$\downarrow$\\ \hline
Xian  \cite{xian2020structure}         &$7.72$  & \multicolumn{1}{l|}{$9.68$}         &$5.00$ & \multicolumn{1}{l|}{$44.77$}   &$0.301$    \\
MegaDepth  \cite{li2018megadepth}    & $4.09$         & \multicolumn{1}{l|}{$8.28$} & $7.04$ &\multicolumn{1}{l|}{ $33.03$} &$0.20$ \\
MiDaS  \cite{Ranftl2020}        &$1.91$ & \multicolumn{1}{l|}{$5.72$} &$3.43$ & \multicolumn{1}{l|}{$12.78$}    &$0.104$    \\

3D Ken Burns  \cite{Niklaus_TOG_2019}   & $2.02$ & \multicolumn{1}{l|}{$\textbf{5.44}$} & $\underline{2.19}$   & \multicolumn{1}{l|}{$\underline{10.24}$} & $\underline{0.097}$  \\
\hline \hline
Ours\tnote{\dag} \, w/o PWN  & $2.05$ & \multicolumn{1}{l|}{$6.10$}  & $3.91$  & \multicolumn{1}{l|}{$13.47$} & $0.106$  \\
Ours\tnote{\dag}    & $\underline{1.91}$ & \multicolumn{1}{l|}{$\underline{5.70}$}  & $2.95$  & \multicolumn{1}{l|}{$11.59$} & $0.101$       \\
Ours Full   & $\textbf{1.90}$ & \multicolumn{1}{l|}{$5.73$}  & $\textbf{2.0}$  & \multicolumn{1}{l|}{$\textbf{7.41}$}  & $\textbf{0.079}$      \\
\bottomrule
\end{tabular}}
\end{threeparttable}
\vspace{0.4em}
\caption{Quantitative comparison of the quality of depth boundaries (DBE) and planes (PE) on the iBims-1 dataset. We use $^\dag$ to indicate when a method was trained on the small training subset.\label{Tab: cmp of edges and planes}}
\vspace{-2em}
\end{table}

\subsection{Monocular Depth Estimation Module}
In this section, we conduct several experiments to demonstrate the effectiveness of our depth prediction method, including a comparison with state-of-the-art methods, a comparison of our proposed image-level normalized regression loss with other methods, and an analysis of the effectiveness of our pair-wise normal regression loss. 

\begin{table*}[t]
\setlength{\tabcolsep}{2pt}
\resizebox{\linewidth}{!}{%
\begin{tabular}{ r | l |ll|ll|ll|ll|ll|ll|ll|l}
\toprule[1pt]
\multirow{2}{*}{Method} & \multirow{2}{*}{Backbone} & OASIS & YT3D & \multicolumn{2}{c|}{NYU} & \multicolumn{2}{c|}{KITTI} & \multicolumn{2}{c|}{DIODE} & \multicolumn{2}{c|}{ScanNet} & \multicolumn{2}{c|}{ETH3D} & \multicolumn{2}{c|}{Sintel} & \multirow{2}{*}{Rank$\downarrow$} \\
&   &\multicolumn{2}{c|}{WHDR$\downarrow$}    & AbsRel$\downarrow$    & $\delta_{1}\uparrow$     & AbsRel$\downarrow$      & $\delta_{1}\uparrow$      & AbsRel$\downarrow$      & $\delta_{1}\uparrow$      &AbsRel$\downarrow$      & $\delta_{1}\uparrow$       &AbsRel$\downarrow$     & $\delta_{1}\uparrow$      & AbsRel$\downarrow$       & $\delta_{1}\uparrow$      &                       \\ \hline
OASIS  \cite{chen2020oasis}  &ResNet50  & $32.7$ &$27.0$ &$21.9$ &$66.8$ &$31.7$ & $43.7$ &  $48.4$ &$53.4$ &$19.8$ &$69.7$ &$29.2$ &$59.5$ &$60.2$ &$42.9$  &  $6.7$\\ 
MegaDepth  \cite{li2018megadepth}& Hourglass &$33.5$  &$26.7$  &$19.4$& $71.4$ &$20.1$ &$66.3$ &$39.1$ &$61.5$ &$19.0$ &$71
.2$ &$26.0$  &$64.3$ &$39.8$ &$52.7$  &$6.7$ \\
Xian \etal \cite{xian2020structure} &ResNet50 &$31.6$ &$23.0$ &$16.6$ &$77.2$ & $27.0$  & $52.9$ &$42.5$ &$61.8$ &$17.4$ &$75.9$ &$27.3$ &$63.0$ &$52.6$ &$50.9$ & $6.7$    \\
WSVD  \cite{wang2019web} &ResNet50 &$34.8$  &$24.8$  &$22.6$ &$65.0$ &$24.4$ &$60.2$ &$35.8$ &$63.8$ &$18.9$ &$71.4$ &$26.1$ &$61.9$ &$35.9$  & $54.5$ &$6.6$ \\
Chen \etal \cite{chen2019learning} &ResNet50  &$33.6$ &$20.9$  &  $16.6$ & $77.3$ & $32.7$ & $51.2$ &$37.9$ & $66.0$ & $16.5$ & $76.7$  &$23.7$ &$67.2$ &$38.4$ & $57.4$ &$5.6$    \\
DiverseDepth  \cite{yin2020diversedepth,yin2021virtual}&ResNeXt50 &$30.9$ &$21.2$ &$11.7$ &$87.5$ &$19.0$ &$70.4$ &$37.6$ &$63.1$ &$10.8$ &$88.2$ &$22.8$ &$69.4$ &$38.6$ &$58.7$  &$4.4$  \\
MiDaS  \cite{Ranftl2020}&ResNeXt101 &$\underline{29.5}$ &$19.9$ &$11.1$ &$88.5$ &$23.6$ &$63.0$ &$33.2$ &$71.5$ &$11.1$ &$88.6$  & $18.4$ &$75.2$ &$40.5$ &$60.6$ & $3.5$ \\
\hline
Ours &ResNet50 &$30.2$ &$\underline{19.5}$  &$\underline{9.1}$  &$\underline{91.4}$  &$\textbf{14.3}$ &$\textbf{80.0}$ &$\underline{28.7}$ &$\underline{75.1}$ &$\underline{9.6}$ &$\underline{90.8}$ &$\underline{18.4}$ &$\underline{75.8}$ &$\underline{34.4}$ &$\underline{62.4}$ &$\underline{1.9}$   \\     
Ours &ResNeXt101 &$\textbf{28.3}$ &$\textbf{19.2}$  &$\textbf{9.0}$  &$\textbf{91.6}$  &$\underline{14.9}$ &$\underline{78.4}$ &$\textbf{27.1}$ &$\textbf{76.6}$ &$\textbf{9.5}$ &$\textbf{91.2}$ &$\textbf{17.1}$ &$\textbf{77.7}$ &$\textbf{31.9}$ &$\textbf{65.9}$ &$\textbf{1.1}$  
\\ \toprule[1pt]
\end{tabular}}
\caption{\textbf{Quantitative comparison} of our depth prediction with state-of-the-art methods on eight zero-shot (unseen during training) datasets. Our method achieves better performance than existing state-of-the-art methods across all test datasets. \label{Tab: Cmp with SOTA}}
\end{table*}

\noindent\textbf{Comparison with state-of-the-art methods.}
In this comparison, we test on datasets unseen during training.
We compare with methods that have been shown to best generalize to in-the-wild scenes. Their results are obtained by running the publicly released code. Each method is trained on its own proposed datasets.
When comparing the AbsRel$\downarrow$ error, we follow Ranftl  \cite{Ranftl2020} and align the scale and shift before the evaluation. 
The results are shown in Table~\ref{Tab: Cmp with SOTA}. Our method outperforms prior works, and using a larger ResNeXt101 backbone further improves the results. Some qualitative comparisons 
are shown 
in Fig.~\ref{Fig: cmp of SOTA. }.

 \noindent\textbf{Pair-wise normal loss.} 
 To evaluate the quality of our full method and dataset on edges and planes, we compare our depth model with previous state-of-the-art methods on the iBims-1 dataset.
 In addition, we evaluate the effect of our proposed pair-wise normal (PWN) loss through an ablation study. As training on our full dataset is computationally demanding, we perform this ablation on the small training subset.
 The results are shown in  Table~\ref{Tab: cmp of edges and planes}. 
 We can see that our full method outperforms prior methods for this task.
 In addition, under the same settings, both edges and planes are improved by adding the PWN loss. 
 We further show a qualitative comparison of depths and reconstructed point clouds in Fig.~\ref{Fig: cmp of depth with PWN. } and Fig.~\ref{Fig: cmp of pcd with PWN. } respectively. We can see that the edges in depths are more accurate and sharper than those without PWN supervision, and the reconstructed point clouds have much fewer distortions.

\noindent\textbf{Image-level normalized regression loss.}
To show the effectiveness of our proposed image-level normalized regression (ILNR) loss, we compare it with the scale-shift invariant loss (SSMAE)  \cite{Ranftl2020} and the scale-invariant multi-scale gradient loss  \cite{wang2019web}. 
Each of these methods is trained on the small training subset to limit the computational overhead, and comparisons are made to datasets that are unseen during training. 
All models have been trained for $50$ epochs, and we have verified that all models are fully converged by then. 
The quantitative comparison is shown in  Table~\ref{ Table cmp of normalization loss}, where we can see an improvement of ILNR over other scale-shift invariant losses. 
Furthermore, we also analyze different options for normalization, including image-level Min-Max (MinMax) normalization and image-level median absolute deviation (MAD) normalization, and found that our proposed loss performs a bit better. We further investigate the effectiveness of `Tanh' term in our ILNR. It can slightly improve the performance. 

\begin{table}[t]
\resizebox{\linewidth}{!}{%
\begin{tabular}{l|c|cccc}
\toprule[1pt]
Method     & RedWeb & NYU & KITTI & ScanNet & DIODE \\
           & WHDR$\downarrow$   & \multicolumn{4}{c}{AbsRel$\downarrow$}    \\ \hline
SMSG  \cite{wang2019web} & $19.1$ & $15.6$ &$16.3$ &$13.7$ & $36.5$\\
SSMAE  \cite{Ranftl2020} & $19.2$ & $14.4$ &$18.2$ &$13.3$ & $34.4$\\\hline  \hline
MinMax  & $19.1$ &$15.0$ & $17.1$ & $13.3$ & $46.1$ \\
MAD   &$18.8$  &$14.8$ &$17.4$ &$12.5$  & $34.6$  \\ \hline  \hline
ILNR-W/o Tanh  &$\textbf{18.6}$ &$14.3$ &$16.3$ &$12.5$ & $34.5$ \\
ILNR  &$18.7$ & $\textbf{13.9}$ & $\textbf{16.1}$ & $\textbf{12.3}$ &$\textbf{34.2}$ \\ \toprule[1pt]
\end{tabular}}
\caption{Quantitative comparison of different losses on zero-shot generalization to $5$ datasets unseen during training.\label{ Table cmp of normalization loss}}
\vspace{-1 em}
\end{table} 

\noindent\textbf{Comparison of depth prediction on people.}
Li~\etal  \cite{li2019learning} propose the first work to solve the depth prediction of moving people. Apart from RGB image, they propose to input the background depth which is obtained by the structure-from-motion method, and the mask of humans as guidance (see Li-IFCM and Li-IDCM in Table~\ref{Tab: people cmp on TUM}) to predict the high-quality depth of moving people.  
In comparison, 
our method only takes a single %
RGB image. Following  \cite{li2019learning}, we conduct the comparison on the  TUM-RGBD  \cite{sturm12iros} dataset. The quantitative comparison illustrated in  Table~\ref{Tab: people cmp on TUM} shows that our method can achieve comparable performance with them. On humans, our depth is more accurate than other methods. Moreover, the visual results are illustrated in Fig.~\ref{Fig: people cmp on TUM. }. We can see that our predicted depths have fewer artifacts and sharper edges than Li \etal \cite{li2019learning} and DiverseDepth  \cite{yin2020diversedepth, yin2021virtual}.

\begin{table}[]
\resizebox{\linewidth}{!}{%
\begin{tabular}{lllll}
\toprule[1pt]
\multicolumn{1}{l|}{Method}  &  \multicolumn{1}{c}{Si-hum$\downarrow$} & \multicolumn{1}{l}{Si-env$\downarrow$} & \multicolumn{1}{l}{Si-RMS$\downarrow$} & AbsRel$\downarrow$ \\ \hline 
\multicolumn{1}{l|}{DeMoN  \cite{ummenhofer2017demon}}  &$0.360$  &$0.302$ &$0.866$ &$0.220$   \\
\multicolumn{1}{l|}{Li-I  \cite{li2019learning}} &$0.294$  &$0.334$ &$0.318$ &$0.204$   \\
\multicolumn{1}{l|}{Li-IFCM  \cite{li2019learning}} &$0.302$ &$0.330$ &$0.316$ &$0.206$     \\
\multicolumn{1}{l|}{Li-IDCM  \cite{li2019learning}} &$0.293$ &$\textbf{0.238}$ &$0.272$ &$\textbf{0.147}$ \\
\multicolumn{1}{l|}{DiverseDepth  \cite{yin2020diversedepth}}&$\underline{0.272}$ &$0.270$ &$\underline{0.272}$  &$0.192$ \\ \hline
\multicolumn{1}{l|}{Ours}&$\textbf{0.258}$ &$\underline{0.247}$ &$\textbf{0.251}$  &$\underline{0.175}$ \\ 
\toprule[1pt]
\end{tabular}}
\caption{%
Comparison of the foreground people on the TUM-RGBD datasets. Our overall performance is comparable with previous methods, while our depths are more accurate on foreground people.
Note that \cite{li2019learning} needs extra input such as the semantic human masks.  
\label{Tab: people cmp on TUM}}
\vspace{-1 em}
\end{table}

\iffalse
\begin{figure}[t]
\centering
\includegraphics[width=\linewidth]{./Figs/TUM_cmp.pdf}
\caption{
\textbf{Qualitative comparison} on the TUM-RGBD dataset. Following Li \etal \cite{li2019learning}, we compare the depth of moving people on
the 
TUM-RGBD dataset.
%
%
}
\label{Fig: people cmp on TUM. }
%
\end{figure}
\fi

\begin{figure*}[t]
\centering
\includegraphics[width=\linewidth]{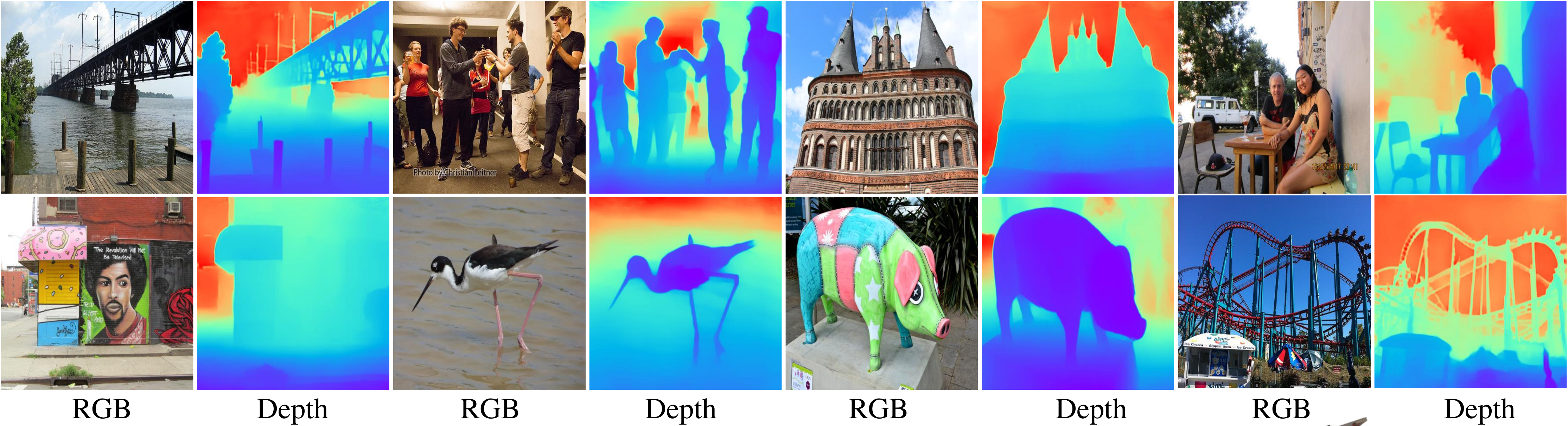}
\includegraphics[width=\linewidth]{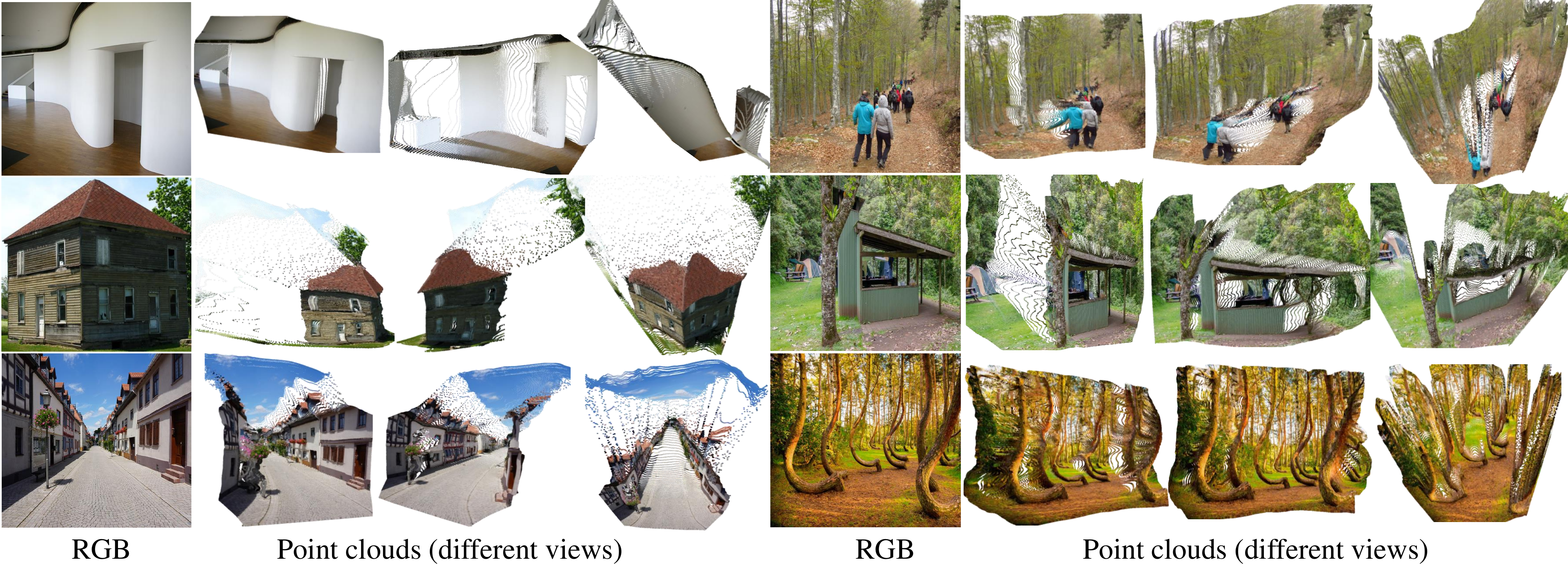}
\caption{\textbf{Qualitative %
results on some 
in-the-wild scenes. }
The reconstructed point clouds and depth maps of some in-the-wild scenes are illustrated. }
\label{Fig: vis in-the-wild pcd and depth . }
\end{figure*}

\noindent\textbf{Additional qualitative results on in-the-wild scenes.}
Fig.~\ref{Fig: vis in-the-wild pcd and depth . } demonstrates more in-the-wild scenes examples. We can see that the predicted depths %
exhibit 
fine details on the edges. %
Furthermore, we %
show 
reconstructed point clouds.

\subsection{Depth Completion}

\begin{table*}[]
\centering
\footnotesize
\setlength{\tabcolsep}{3pt}
\begin{tabular}{l|ccc|ccc|ccc}
\toprule[1pt]
\multirow{3}{*}{Methods}
                  & \multicolumn{3}{c|}{NYU}                                                                                          & \multicolumn{3}{c|}{ScanNet}                                                                                      & \multicolumn{3}{c}{DIODE}                                                                                        \\ %
                  \cline{2-10}
                 
                  & \multicolumn{3}{c|}{AbsRel$\downarrow$}                                                                                       & \multicolumn{3}{c|}{AbsRel$\downarrow$}                                                                                       & \multicolumn{3}{c}{AbsRel$\downarrow$} \\
                  & \multicolumn{1}{c}{Unp. FOV} & \multicolumn{1}{c}{Sparse ToF} & \multicolumn{1}{c|}{Short Range} & \multicolumn{1}{c}{Unp. FOV} & \multicolumn{1}{c}{Sparse ToF} & \multicolumn{1}{c|}{Short Range}  & \multicolumn{1}{c}{Unp. FOV} & \multicolumn{1}{c}{Sparse ToF} & \multicolumn{1}{c}{Short Range}  \\ \hline
NLSP~\cite{park2020non} &$0.150$  &$0.190$ &$0.114$ & $0.716$ &$1.413$   &$0.202$ &$6.684$ &$11.370$ &$1.005$ \\
Senushkin~\etal~\cite{senushkin2020decoder}&$0.224$ &$0.615$  &$0.093$  &$0.255$ &$0.793$  &$0.166$ &$0.687$ &$6.120$     &$0.623$  \\
Ours-Baseline &$\textbf{0.046}$ &$\textbf{0.018}$   &$\textbf{0.041}$ &$\textbf{0.049}$ &$\textbf{0.022}$   &$\textbf{0.047}$  &$\textbf{0.150}$ &$\textbf{0.143}$  &$\textbf{0.144}$                         \\ 
Ours   &$\textbf{0.031}$ &$\textbf{0.013}$  &$\textbf{0.030}$ &$\textbf{0.028}$ &$\textbf{0.014}$ &$\textbf{0.037}$ &$\textbf{0.139}$ &$\textbf{0.111}$  &$\textbf{0.137}$                 \\ \toprule[1pt]
\end{tabular}%
\vspace{-1 em}
\caption{Comparison of our method with state-of-the-art methods on zero-shot test datasets. We create $3$ different sparse depth types for evaluation. It is clear that our method has better          generalization        than previous methods on unseen data and unseen sparsity patterns. 
\label{Tab: Generalization cmp.}}
\vspace{-0.3 em}
\end{table*}

In this section, we conduct several experiments to %
report the effectiveness of our method for depth completion. To %
show 
the          generalization        of our method, we conduct the zero-shot testing on a few benchmark datasets. Note that we only train a single model to solve different sparse depth situations,  %
while 
previous methods  \cite{park2020non, senushkin2020decoder} train different models for different sparse patterns. 

\noindent\textbf{Comparison with state-of-the-art 
depth completion methods.} We test on standard benchmarks, NYU  \cite{silberman2012indoor} and Matterport3D  \cite{Matterport3D}.  Note that our models %
were not trained on these datasets.
Two benchmarks have different types of sparse patterns: On NYU, the sparse depth only has 500 valid pixels, while Matterport3D provides the incomplete sensor-captured depth map. We include a baseline method (Ours-baseline) where the model does not make use of depth guidance, that is, it directly predicts the complete depth from RGB and sparse depth without the guidance map. `Ours' input another guidance map, which is from our DPM.

\begin{figure*}[t]
\centering
\includegraphics[width=\linewidth]{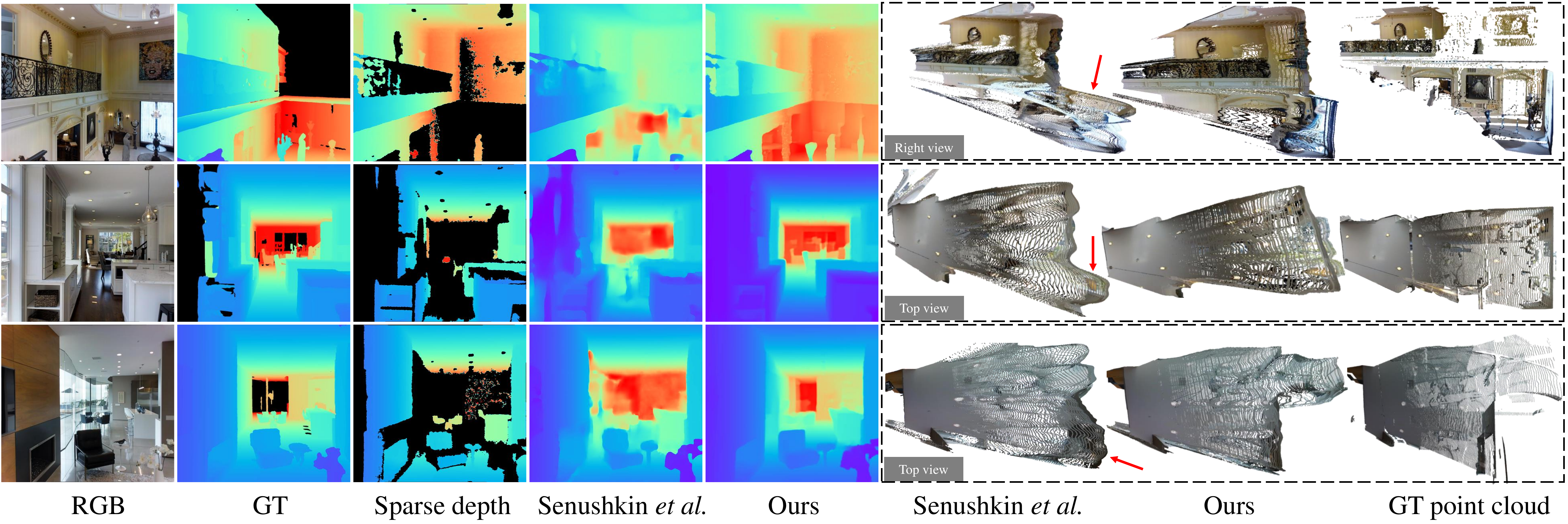}
\caption{\textbf{Qualitative comparison of 
the 
depth and reconstructed 3D shapes}. Our 
metric depths %
show
finer details. The reconstructed 3D shape is more accurate than previous methods.}
\label{Fig: cmp of depth and pcd on matterport3d. }
\end{figure*}

\begin{table}[]
\centering
\footnotesize
\setlength{\tabcolsep}{5pt}
\begin{tabular}{l|lllll}
\toprule[1pt]
Methods                & RMSE$\downarrow$ & AbsRel$\downarrow$ & $\delta_{1}\uparrow$     & $\delta_{2}\uparrow$      & $\delta_{3}\uparrow$       \\ \hline
S2D~\cite{ma2018sparse}        & $0.230$   & $0.044$   & $97.1$ & $99.4$ & $99.8$  \\
S2D+SPN~\cite{liu2017learning} & $0.172$   & $0.031$   & $98.3$ & $99.7$ & $99.9$  \\
DepthCoeff~\cite{imran2019depth}   & $0.118$   & $0.013$   & $99.4$ & $99.9$ & -     \\
CSPN~\cite{cheng2018depth}         & $0.117$   & $0.016$   & $99.2$ & $99.9$ & $100.0$ \\
DeepLiDAR~\cite{qiu2019deeplidar}    & $0.115$   & $0.022$   & $99.3$ & $99.9$ & $100.0$ \\
DepthNormal~\cite{xu2019depth} & $0.112$   & $0.018$   & $99.5$ & $99.9$ & $100.0$ \\
NLSP~\cite{park2020non}                 & $0.092$   & $\textbf{0.012}$   & $99.6$ & $99.9$ & $100.0$\\ 
Lee~\etal~\cite{lee2021depth} &$0.104$ &$0.014$ &$99.4$ &$99.9$ &$100.0$ \\ 
\hline
Ours-baseline &$0.210$  &$0.036$ &$98.4$ &$99.6$ &$99.9$       \\
Ours  &$0.183$  &$0.022$ &$98.7$ &$99.7$ &$99.9$       \\\toprule[1pt]
\end{tabular}
\vspace{-1 em}
\caption{Depth completion results on the NYU dataset. Following ~\cite{ park2020non}, we uniformly sample 500 points from ground truth as the sparse depth. Our method is \emph{not trained on NYU} but is comparable with state-of-the-art methods show here that \emph{are} trained on NYU. With guidance map, the performance is better.}\label{Tab: DepComplete Cmp with SOTA on NYU}
\vspace{-1 em}
\end{table}

Table~\ref{Tab: DepComplete Cmp with SOTA on NYU} reports the 
results on NYU. Our method performance is close to these state-of-the-art methods, and better than the baseline. Similarly, Table~\ref{Tab: DepComplete Cmp with SOTA on Matterport3D} shows the comparison on the Matterport3D dataset. %
Ours can outperform previous methods on some metrics, and are better than the baseline. Note that we do not fine-tune our model on the target Matterport3D dataset.

\begin{table}[t]
\centering
\footnotesize
\setlength{\tabcolsep}{2pt}
\begin{tabular}{l|lllllll}
\toprule[1pt]
Methods           & RMSE$\downarrow$  & MAE$\downarrow$   & $\delta_{1.05}\uparrow$ & $\delta_{1.1}\uparrow$  & $\delta_{1}\uparrow$  & $\delta_{2}\uparrow$ & $\delta_{3}\uparrow$  \\ \hline
Huang~\etal~\cite{huang2019indoor} & $1.092$ & $0.342$ & $66.1$ & $75.0$ & $85.0$ & $91.1$ & $93.6$ \\
Zhang~\etal~\cite{zhang2018deepdepth}& $1.316$ & $0.461$ & $65.7$ & $70.8$ & $78.1$ & $85.1$ & $88.8$    \\
Senushkin~\etal~\cite{senushkin2020decoder}& $\textbf{1.03}$ & $\textbf{0.299}$ & $\textbf{71.9}$ & $\textbf{80.5}$ & $\textbf{89.0}$ & $\textbf{93.2}$  & $95.0$  \\ \hline
Ours-baseline & $2.35$ & $0.574$ & $68.9$ & $78.6$ & $86.1$ & $91.5$ & $96.0$                             \\
Ours   & $\textbf{1.03}$ & $\underline{0.320}$ & $\underline{71.2}$ & $\underline{79.0}$ & $\underline{87.1}$ & $\underline{93.1}$ & $\textbf{96.0}$                           \\ \toprule[1pt]
\end{tabular}
\vspace{-1 em}
\caption{Depth completion results on the Matterport3D dataset. Our method is \emph{not trained on Matterport3D} but is comparable with state-of-the-art methods that are trained on Matterport3D. RMSE and MAE are given in meters. With the guidance map, the performance is improved. }\label{Tab: DepComplete Cmp with SOTA on Matterport3D}
\vspace{-1 em}
\end{table}

\noindent\textbf{Illustration of metric reconstruction.}
We show some visual comparisons of reconstructed metric shape in Fig.~\ref{Fig: cmp of depth and pcd on matterport3d. }. Note that `ours' employs PCM to predict the focal length, while others use the ground truth focal length. Although the state-of-the-art depth completion method can achieve better quantitative performance  than ours, our method---supervised by the geometric loss---can reconstruct a more accurate scene structure. It is clear that our reconstructed walls are flatter than \cite{senushkin2020decoder}. 

\noindent\textbf{Effectiveness of completing sparse depth.}
To evaluate the robustness of our method to noisy sparse inputs, we use COLMAP~\cite{schoenberger2016mvs} to densely reconstruct $16$ scenes of NYU. We use the ground truth to remove some outliers to obtain the sparse depth as inputs. There are over $4000$ images for evaluation. Results are shown in ~\ref{Tab: Complete Noisy cmp.}. We mainly compare with NLSPN~\cite{park2020non} and Senushkin~\etal~\cite{senushkin2020decoder}, which have achieved the state-of-the-art performance on benchmarks. Our
method can achieve better accuracy than existing methods. Comparing with our baseline, using the guidance map further boosts performance. Some qualitative comparisons are shown in Fig.~\ref{Fig: complete noisy depth visually.}, we can see that our completed depths have much less outliers and noise (see the wall).

\begin{table}[]
 \resizebox{\linewidth}{!}{
\begin{tabular}{l|cc|cc}
\toprule[1pt]
Metrics
           &\begin{tabular}[c]{@{}c@{}}NLSPN\\ \cite{park2020non}\end{tabular}  & \begin{tabular}[c]{@{}c@{}}Senushkin\etal\\ ~\cite{senushkin2020decoder}\end{tabular} & \begin{tabular}[c]{@{}c@{}}Ours\\ (baseline)\end{tabular} & \begin{tabular}[c]{@{}c@{}}Ours\\ (W Guidance)\end{tabular} \\ \hline
AbsRel (\%)$\downarrow$     & 24.9 & 18.26     & 6.41                                                      & \textbf{5.25}                                                        \\
$\delta_1$ (\%) $\uparrow$ & 54.7  & 70.3      & 93.6                                                      & \textbf{95.2}                                                        \\ \toprule[1pt]
\end{tabular}}
\vspace{-0.5 em}
\caption{Comparison on completing noisy sparse depth, showing robustness to noisy input.
\label{Tab: Complete Noisy cmp.}}
\vspace{-2 em}
\end{table}

\noindent\textbf{Generalization to different sparse depth types.}
To demonstrate the robustness of our methods to zero-shot test datasets and some unseen sparsity patterns, we create $3$ sparse depth patterns on $3$ unseen datasets for evaluation. Note that such synthesized sparsity patterns are different from that used in training.
Sparsity patterns are: 1) Unpaired  FOV.  We propose to remove 25\% region along the 4 borders of the ground truth depth as the sparse depth.  2) Sparse ToF. To simulate Time-of-flight sensors captured sparse depth we downsample the ground truth depth map to a low resolution, up-project to the original size, and mask the distant regions to obtain the sparse depth. 3) Short  Range.  We mask the 50\% most distant regions of ground-truth to obtain the incomplete depth. We compare our methods to the state-of-the-art methods of NYU and Matterport3D benchmarks, \textit{i.e.},  NLSPN \cite{park2020non} and Senushkin~\etal~\cite{senushkin2020decoder}. NLSPN method aims to complete the depth with only hundreds of valid points, while Senushkin~\etal~\cite{senushkin2020decoder} method designs to complete contiguous holes. 
We can see that, although NLSP  \cite{park2020non} and Senushkin \etal  \cite{senushkin2020decoder} can achieve state-of-the-art performance on NYU and Matterport3D dataset respectively, they cannot generalize to different types of sparse depth and unseen datasets. %
In 
contrast, our method can achieve comparable performance on different datasets. We believe that our mixed-data training strategy and inputting a guidance map can significantly improve the model's robustness. Some visual comparison is shown in Figure~\ref{Fig: diff sparsity cmp}.

\subsection{Applications}
Our depth predictions can be used to
apply a range of depth-based visual effects such as defocus, view synthesis, and so on. Here we %
show an example of 
using our predicted depth to create a 3D photo. 
We take  the  method of 
\cite{Shih3DP20} to synthesize new views, which takes the single image and our predicted depth map to create new views. Furthermore, we also show an example of using our predicted depth to create the defocus image. Results are shown in Fig.~\ref{Fig: cmp of 3D photo. }.

\subsection{Limitations}
We 
have 
observed a few limitations of our method. Here we %
analyze some failure cases of the depth prediction module and the point cloud module. 

\begin{figure}[t]
\centering
\includegraphics[width=\linewidth]{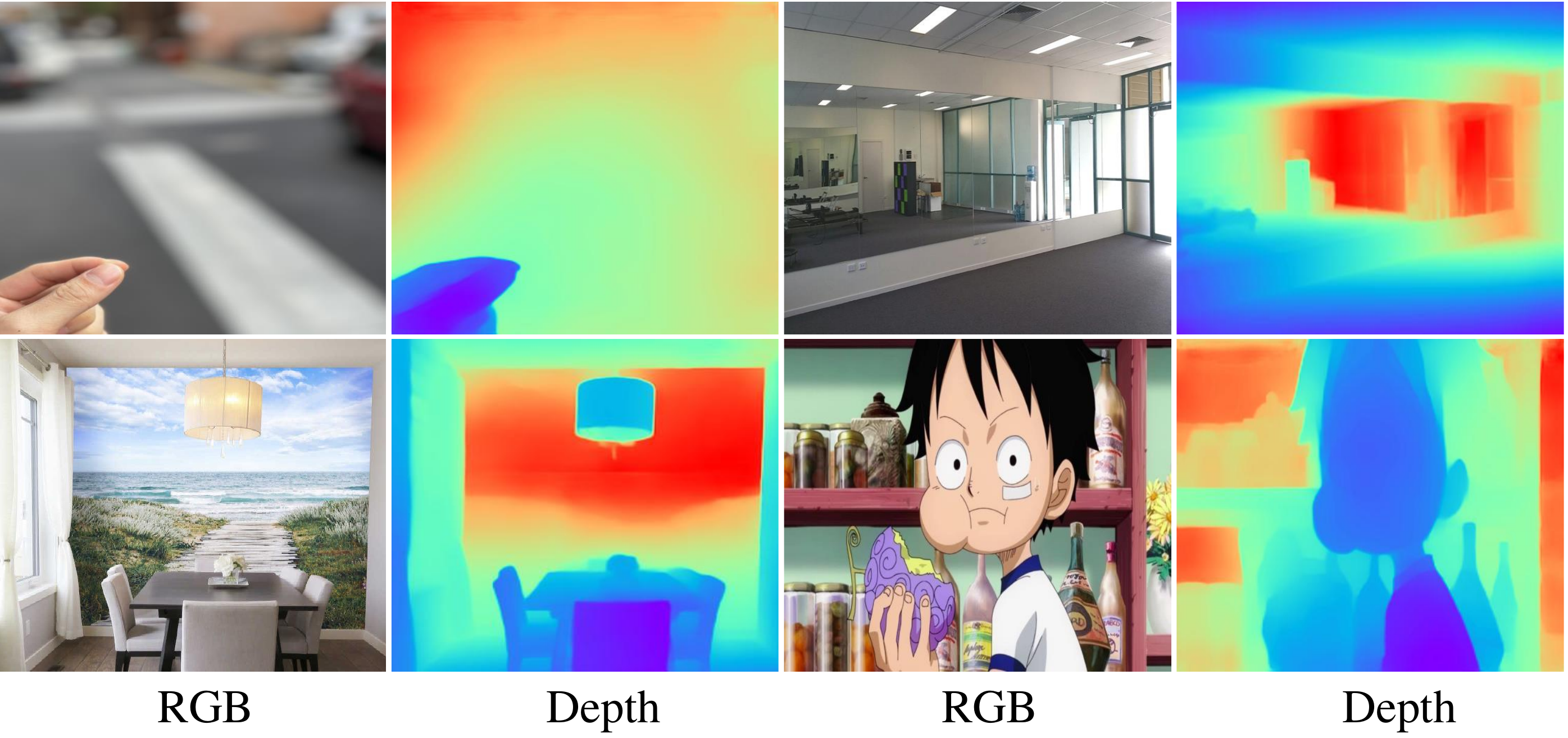}
\caption{\textbf{Failure cases} of the monocular depth prediction module. It is challenging to predict accurate monocular depth for blurry background, mirror, 2D paintings, and cartoon scenes. }
\label{Fig: failed cases of DPM. }
\vspace{-1em}
\end{figure}

\noindent\textbf{Failure cases of DPM.} We %
show 
some typical failure cases of DPM in Fig.~\ref{Fig: failed cases of DPM. }. 1) %
Out of focus. This may be due to the fact that 
our training images are all-in-focus. 
2) %
Paintings (or mirrors) can 
cause ambiguity to the network. 
3) 
Cartoons. Since the %
domain gap exists between cartoons %
and real %
photos,  
the network %
cannot
work well. 
We believe that such problems %
can 
be %
largely solved with more training data.

\begin{figure}[t]
\centering
\includegraphics[width=\linewidth]{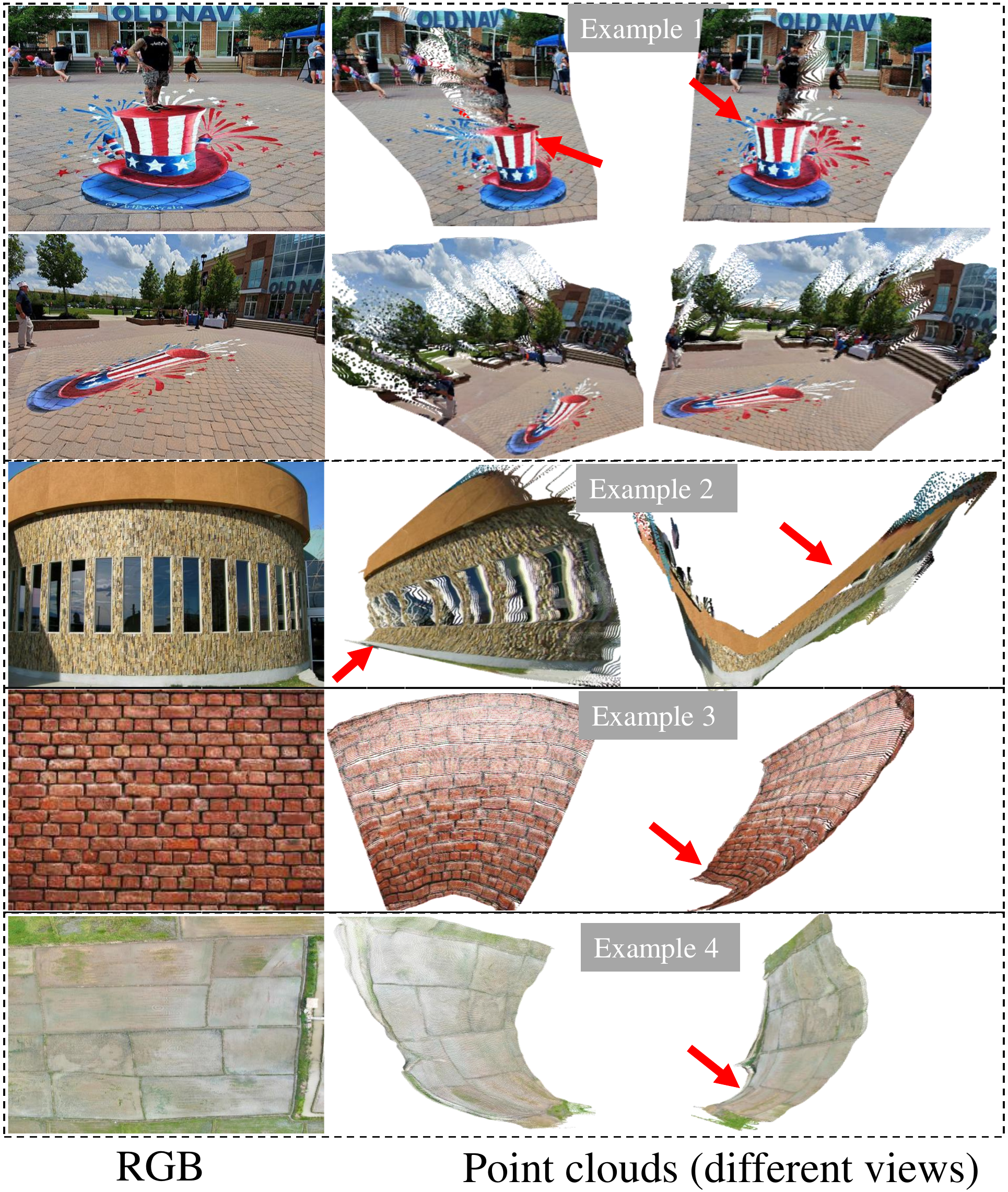}
\caption{\textbf{Failure cases} of recovering 3D shape. Our method fails on 3D painting (`Example 1'), curvy walls (`Example 2'), the scene only with the wall (`Example 3'), and the bird-of-view scene (`Example 4'). In `Example 1', our method is consistent with human visual results. It falsely predicts the `hat' in a 3D shape from a specific point of view, and correctly understands it in another view. The method may also fail on scenarios with less reference background and uncommon view angles.  }
\label{Fig: failed cases of pcd. }
\vspace{-2em}
\end{figure}

\noindent\textbf{Failure cases of PCM.} Our PCM cannot recover accurate focal length or depth shift when the scene does not have enough geometric cues, \eg,
when the whole image is mostly a wall or a sky region (see the `Example 3' in Fig.~\ref{Fig: failed cases of pcd. }). 
The accuracy of our method will also decrease with images taken from uncommon view angles (\eg, top-down) (see `Example 4' in Fig.~\ref{Fig: failed cases of pcd. }) or extreme focal lengths. 
More diverse 3D training data may address these failure cases. 
In addition, our method does not model the effect of radial distortion from the camera and thus the reconstructed scene shape can be distorted in cases with severe radial distortion. 
Studying how to recover the radial distortion parameters using our PCM can be an interesting future direction.

Furthermore, we also obverse that our system will fail on 3D paintings, see the `Example 1' in Fig.~\ref{Fig: failed cases of pcd. }. Our method falsely predicts the 3D painting `hat' in a 3D shape (see the first row), while the problem is relieved a lot when seeing this scene in another view (see the second row). Besides, `Example 2' shows that curvy walls are another difficult case for our method.

\section{Conclusion}

In summary, we have  presented, to our knowledge, the first fully data-driven method that reconstructs 3D scene shapes from single monocular images. 
To recover the shift and focal length for 3D reconstruction, we have proposed to use point cloud networks trained on datasets with known global depth shifts and focal lengths.
This approach has demonstrated  strong generalization capabilities, and we are under the impression that it may be helpful for related depth and 3D reconstruction tasks. 
Our extensive experiments %
verify 
the effectiveness of our scene shape reconstruction method and the superior generalization to unseen data.
\vspace{.3cm}
{\bf 
    Authors' photographs and biographies not available at the time of publication.
} 

{\small
\bibliographystyle{ieeetr}
\bibliography{TPAMI}
}

\clearpage

We provide experiment details and some additional results here.

\appendices

\pagenumbering{roman}
\setcounter{page}{1}

\setcounter{table}{0}
\renewcommand{\thetable}{A\arabic{table}}

\setcounter{figure}{0}
\renewcommand{\thefigure}{A\arabic{figure}}

\section{Details for DCM and PCM training}
\subsection{Networks}

\begin{figure}[ht!]
\centering
\includegraphics[width=\linewidth]{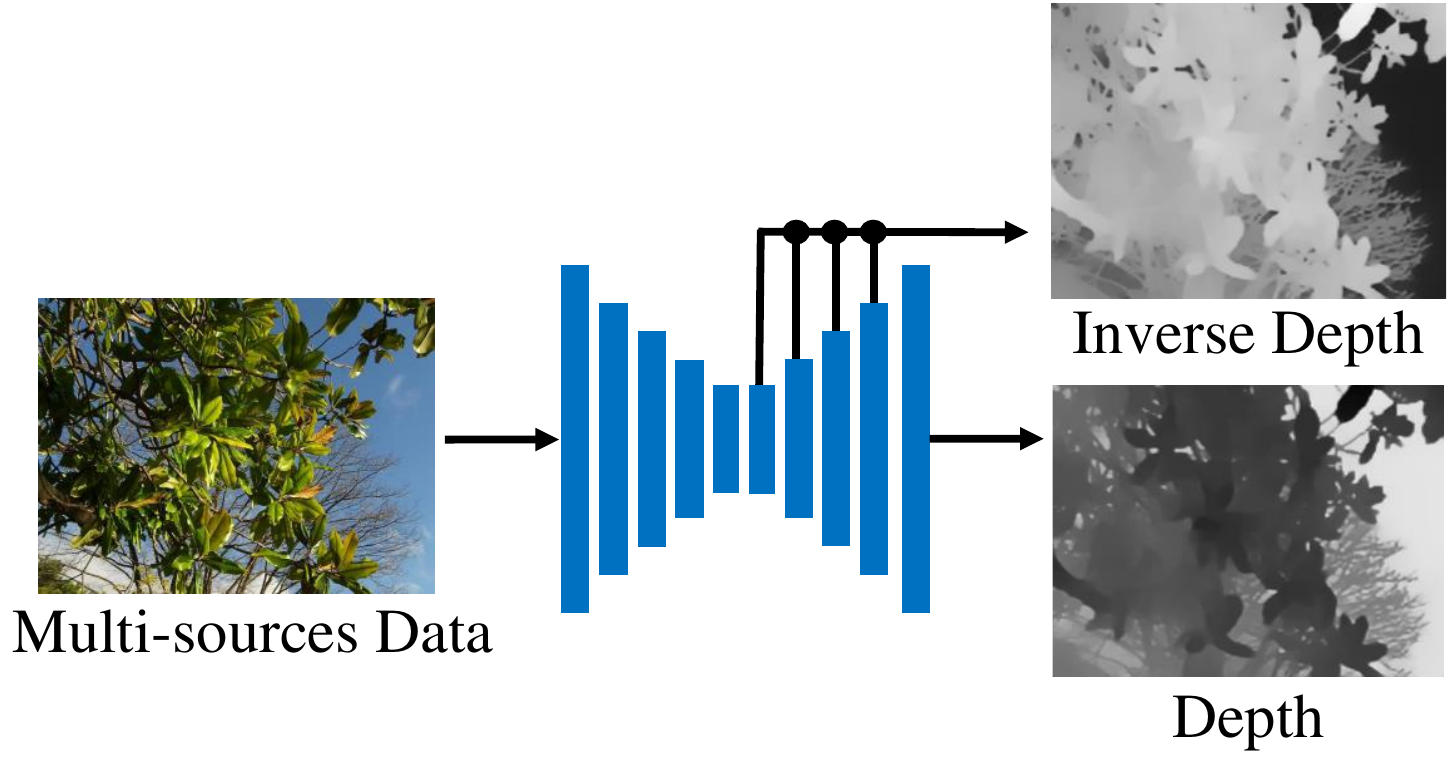}
\caption{The network architecture for the DPM. The network has two output branches. The decoder outputs the depth map, while the auxiliary path outputs the inverse depth. Different losses are enforced on these two branches.}
\label{Fig: network. }
\end{figure}

\subsection{Training}

We use the depth prediction model proposed by Xian~\etal~\cite{xian2020structure}.

We follow~\cite{yin2020diversedepth} and combine the multi-source training data by evenly sampling from all sources per batch. 
As HRWSI and Holopix50K are both web stereo data, we merge them together. Therefore, there are four different data sources, i.e. high-quality Taskonomy, synthetic 3D Ken Burn, middle-quality DIML, and low-quality Holopix50K and HRWSI. For example, if the batch size is $8$, we sample $2$ images from each of the four sources. 
Furthermore, as the ground truth depth quality varies between data sources, we enforce different losses for them. 

For the web-stereo data, such as Holopix50K~\cite{hua2020holopix50k} and HRWSI~\cite{xian2020structure}, as their inverse depths have unknown scale and shift, these inverse depths cannot be used to compute the affine-invariant depth (up to an unknown scale and shift to the metric depth). 
The pixel-wise regression loss and geometry loss cannot be applied for such data. Therefore, during training, we only enforce the ranking loss~\cite{xian2018monocular} on them. 

For the middle-quality calibrated stereo data, such as DIML~\cite{kim2018deep}, we enforce the proposed image-level normalized regression loss, multi-scale gradient loss and ranking loss. As the recovered disparities contain much noise in local regions, enforcing the pair-wise normal regression loss on noisy edges will cause many artifacts. Therefore, we enforce the pair-wise normal regression loss only on planar regions for this data. 

For the high-quality data, such as Taskonomy~\cite{zamir2018taskonomy} and synthetic 3D Ken Burns~\cite{Niklaus_TOG_2019}, accurate edges and planes can be located. Therefore, we apply the pair-wise normal regression loss, ranking loss, and multi-scale gradient loss for this data.

Furthermore, we follow \cite{liu2019training} and add a light-weight auxiliary path on the decoder. 
The auxiliary outputs the inverse depth and the main branch (decoder) outputs the depth. 
For the auxiliary path, we enforce the ranking loss, image-level normalized regression loss in the inverse depth space on all data. 
The network is illustrated in Fig.~\ref{Fig: network. }. 

When training the point cloud network, we follow PVCNN~\cite{liu2019pvcnn} classification training setting but replace the cross-entropy loss with our proposed regression loss. To train the PCM,
we sampled 100K Kinect-captured depth maps from ScanNet, 114K LiDAR-captured depth maps from Taskonomy, and 51K synthetic depth maps from the 3D Ken Burns. We train the network using SGD with a batch size of 40, an initial learning rate of 0.24, and a learning rate decay of 0.1. For parameters specific to PVCNN, such as the
voxel size, we follow the original work.

\section{Details for Depth Completion Training}
\begin{figure*}[ht]
\centering
\includegraphics[width=.85\linewidth]{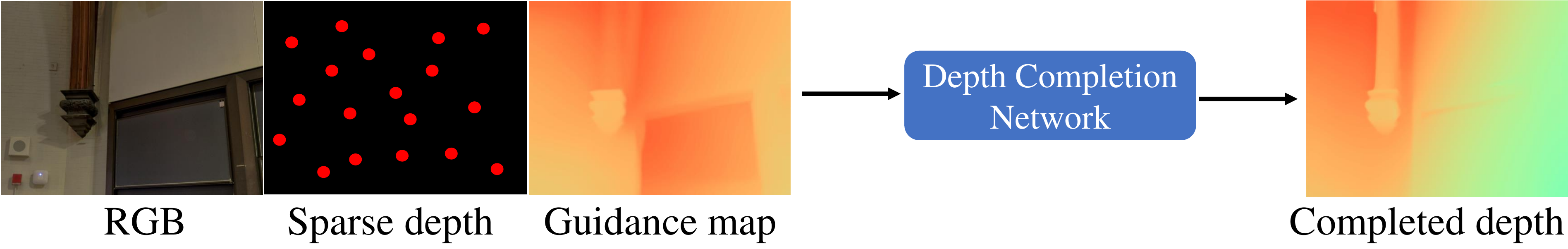}
\caption{
% Qualitative 
\textcolor{black}{\textbf{Framework of depth completion}. We propose to input an RGB image, a sparse depth, and a guidance map for the depth completion. The guidance map is obtained from our monocular depth estimation method.}}
\label{Fig: depth completion framework. }
%\vspace{-1 em}
\end{figure*}

The depth completion module takes an RGB image, a sparse depth map, and our DPM predicted depth as the guidance map. 
The ESANet-R34-NBt1D network~\cite{esanet2020} is employed.  The framework is illustrated in \ref{Fig: depth completion framework. }. During training, we used Taskonomy~\cite{zamir2018taskonomy} and DIML~\cite{kim2018deep} as the training data. The SGD is used for optimization with
an initial learning rate of 0.02. The learning rate is decayed every 40000 iterations with the ratio 0.1. The batch size is 24.

\begin{figure*}[ht]   
\centering
\includegraphics[width=.7\linewidth]{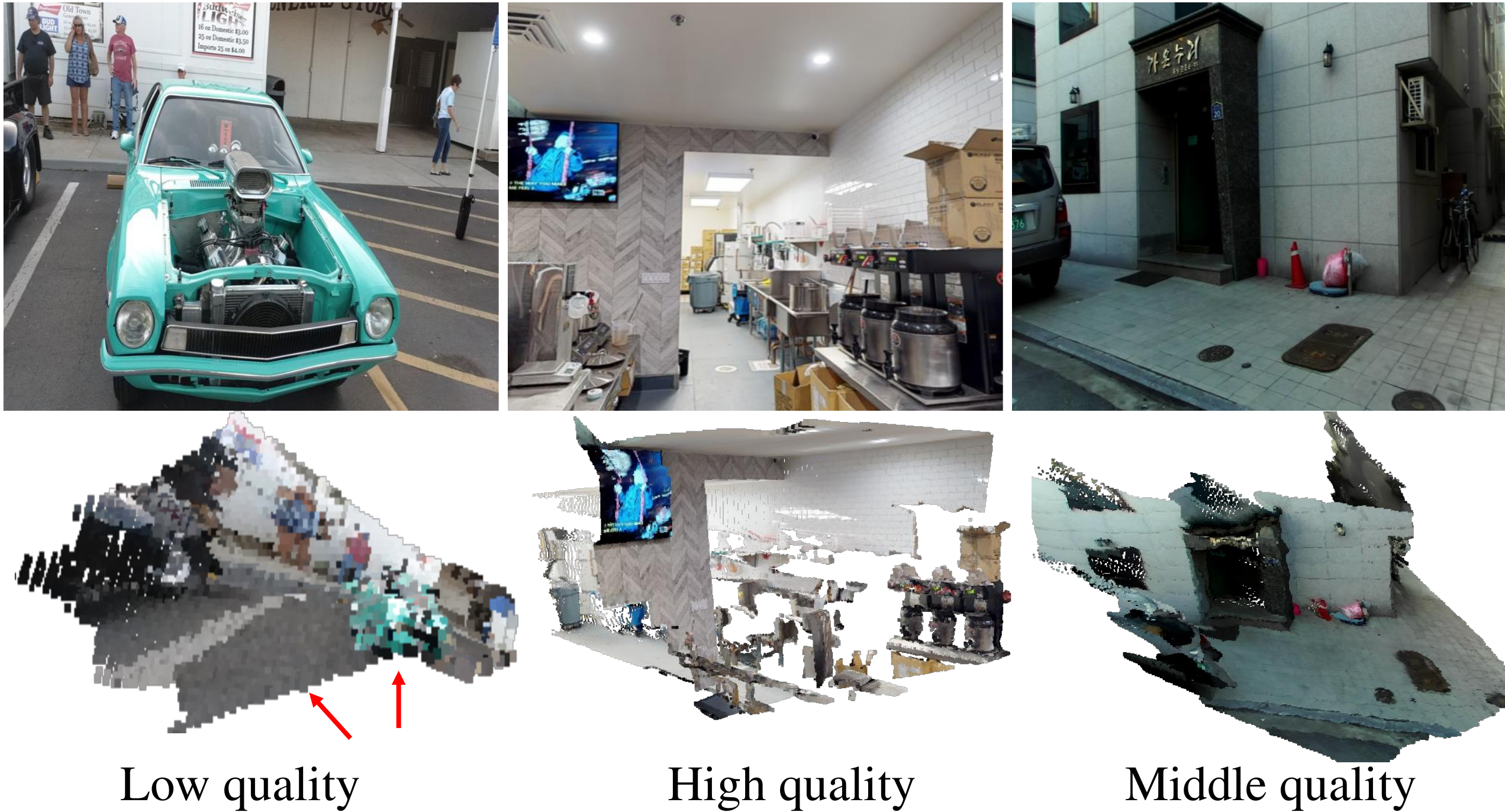}
\caption{\textcolor{black}{Some examples of training data. High-quality and medium-quality data are from Taskonomy and DIML respectively. Low-quality data is from HRWSI, of which the unprojected point cloud 
%has
exhibit 
significant distortion.} }
%\vspace{-0.5em}
\label{Fig: cmp of data.}
\end{figure*}

\begin{table*}[h!]
\centering
\resizebox{.7638\linewidth}{!}{%
\begin{tabular}{l|llll}
\toprule[1pt]
\multirow{1}{*}{Dataset} & \multirow{1}{*}{\# Images} & Scene type & Evaluation metric & Supervision type
 %&  & Type & Metric & Type 
 \\ \hline
NYU & $654$ & Indoor & AbsRel  \& $\delta_{1}$ & Kinect \\ \hline
ScanNet & $700$ & Indoor & AbsRel \& $\delta_{1}$ & Kinect \\ \hline
2D-3D-S & $12256$ & Indoor & LSIV & LiDAR \\ \hline
\multirow{2}{*}{iBims-1} & \multirow{2}{*}{$100$} & \multirow{2}{*}{Indoor} & \multirow{2}{*}{\begin{tabular}[c]{@{}l@{}}AbsRel \&\\ $\varepsilon_\text{PE}$  \&$\varepsilon_\text{DBE}$\end{tabular}} & \multirow{2}{*}{LiDAR} \\
 &  &  &  &  \\ \hline 
KITTI & $652$ & Outdoor & AbsRel \& $\delta_{1}$ & LiDAR \\ \hline
Sintel & $641$ & Outdoor & AbsRel \& $\delta_{1}$ & Synthetic \\ \hline
ETH3D & $431$ & Outdoor & AbsRel \& $\delta_{1}$ & LiDAR \\ \hline 
YouTube3D & $58054$ & In the Wild & WHDR & SfM, Ordinal pairs \\  \hline
\multirow{2}{*}{OASIS} & \multirow{2}{*}{$10000$} & \multirow{2}{*}{In the Wild} & \multirow{2}{*}{\begin{tabular}[c]{@{}l@{}}WHDR %\\
\& LSIV\end{tabular}} & \multirow{2}{*}{\begin{tabular}[c]{@{}l@{}}User clicks, \\ Small patches with GT\end{tabular}} \\
 &  &  &  &  \\ \hline
\multirow{2}{*}{DIODE} & \multirow{2}{*}{$771$} & \multirow{2}{*}{\begin{tabular}[c]{@{}l@{}}Indoor %\\
 \& Outdoor\end{tabular}} & \multirow{2}{*}{\begin{tabular}[c]{@{}l@{}}AbsRel \& %\\
$\delta_{1}$\end{tabular}} & \multirow{2}{*}{LiDAR} \\
 &  &  &  &  \\ \hline
 TUM-RGBD & $1815$ & Indoor & AbsRel \& SiLog & Kinect \\ \toprule[1pt]
\end{tabular}}
\caption{Overview of the test sets in our experiments. \label{Tab: testing data details}}
\vspace{-0.5em}
\end{table*}
\begin{figure*}[ht]
\centering
\includegraphics[width=.807\linewidth]{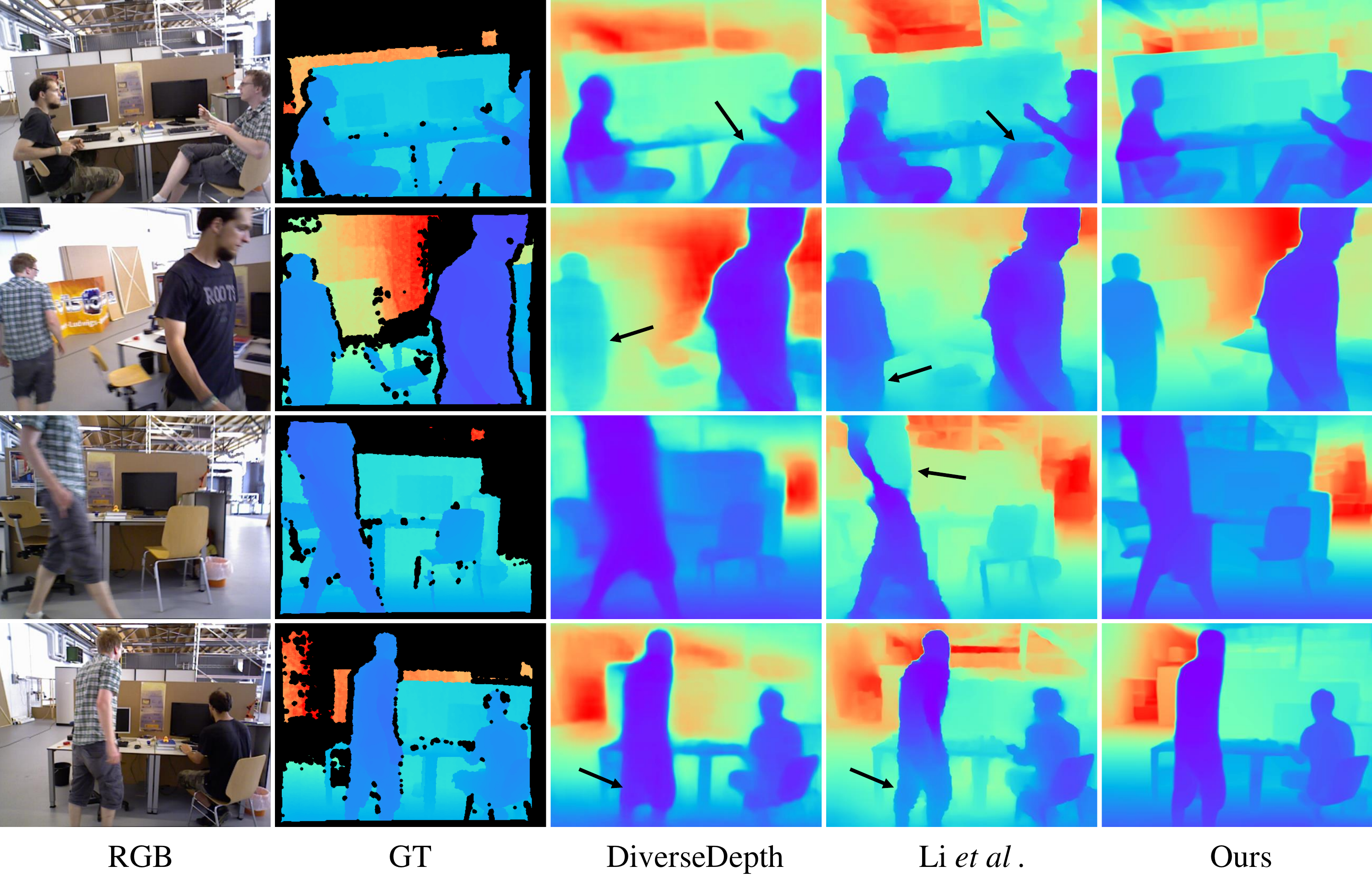}
\caption{
\textbf{Qualitative comparison} on the TUM-RGBD dataset. Following Li \etal \cite{li2019learning}, we compare the depth of moving people on
the 
TUM-RGBD dataset.
%. The black arrow highlight the comparison regions.
%
}
\label{Fig: people cmp on TUM. }
%\vspace{-1em}
\end{figure*}
\begin{figure*}[t]
\centering
\subfloat[]
{
	\includegraphics[trim=0cm 1.48cm 0cm 1.48cm, clip=true, width=0.425\linewidth]{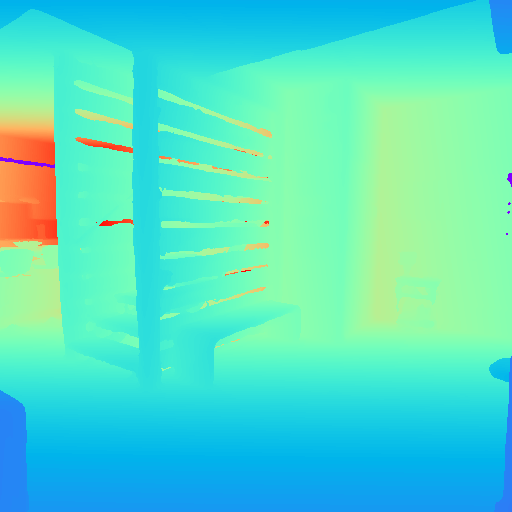}
	\label{Fig: original depth.}
}
\subfloat[]
{
	\includegraphics[trim=0cm 2cm 0cm 2cm, clip=true, width=0.425\linewidth]{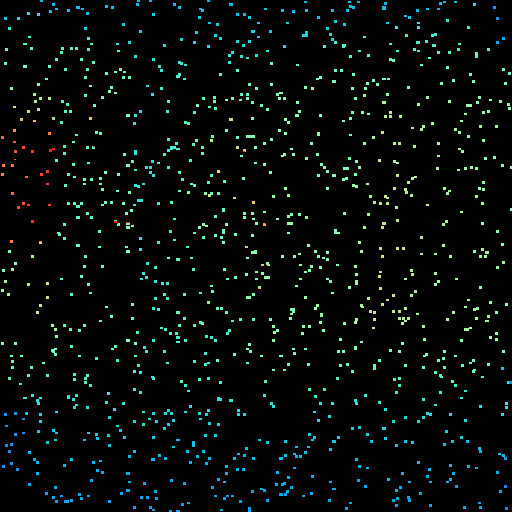}
	\label{Fig: uniform samples.}
}\\
\vspace{0.2cm}
\subfloat[]
{ 
	\includegraphics[trim=0cm 2cm 0cm 2cm, clip=true, width=0.425\linewidth]{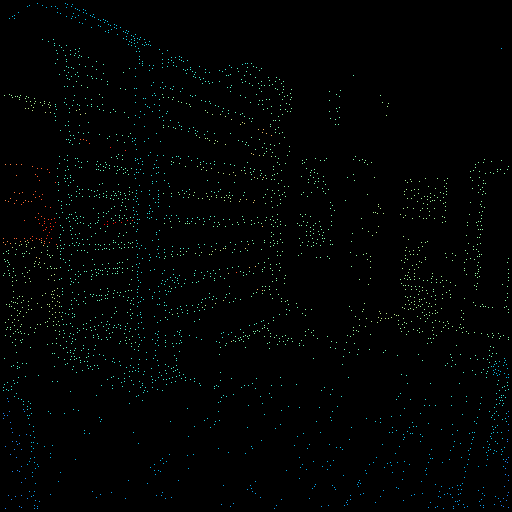} 
	\label{Fig: features.}
}
\subfloat[]
{ 
	\includegraphics[trim=0cm 2cm 0cm 2cm, clip=true, width=0.425\linewidth]{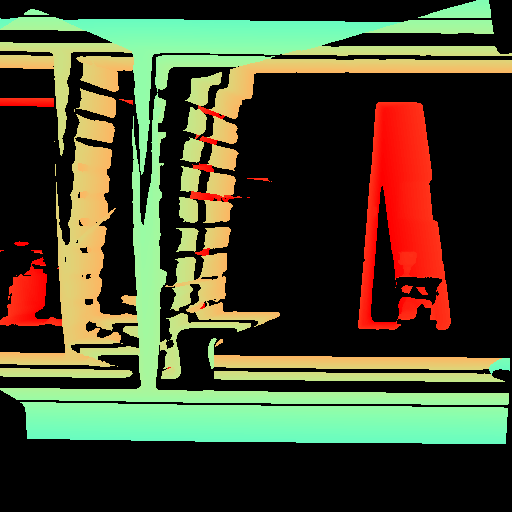} 
	\label{Fig: simulate sensor.}
}
\caption{Visualization of sampled sparse depths. We simulate three different patterns from (a) the dense depth to train models: (b) random uniform sampling, (c) feature point based sampling, and (d) region-based sampling.}
\label{Fig: sdepth patterns}
\vspace{-1 em}
\end{figure*}
\appendices

% Complete noise COLMAP depths
\begin{figure*}[t]
\centering
\includegraphics[width=1\linewidth]{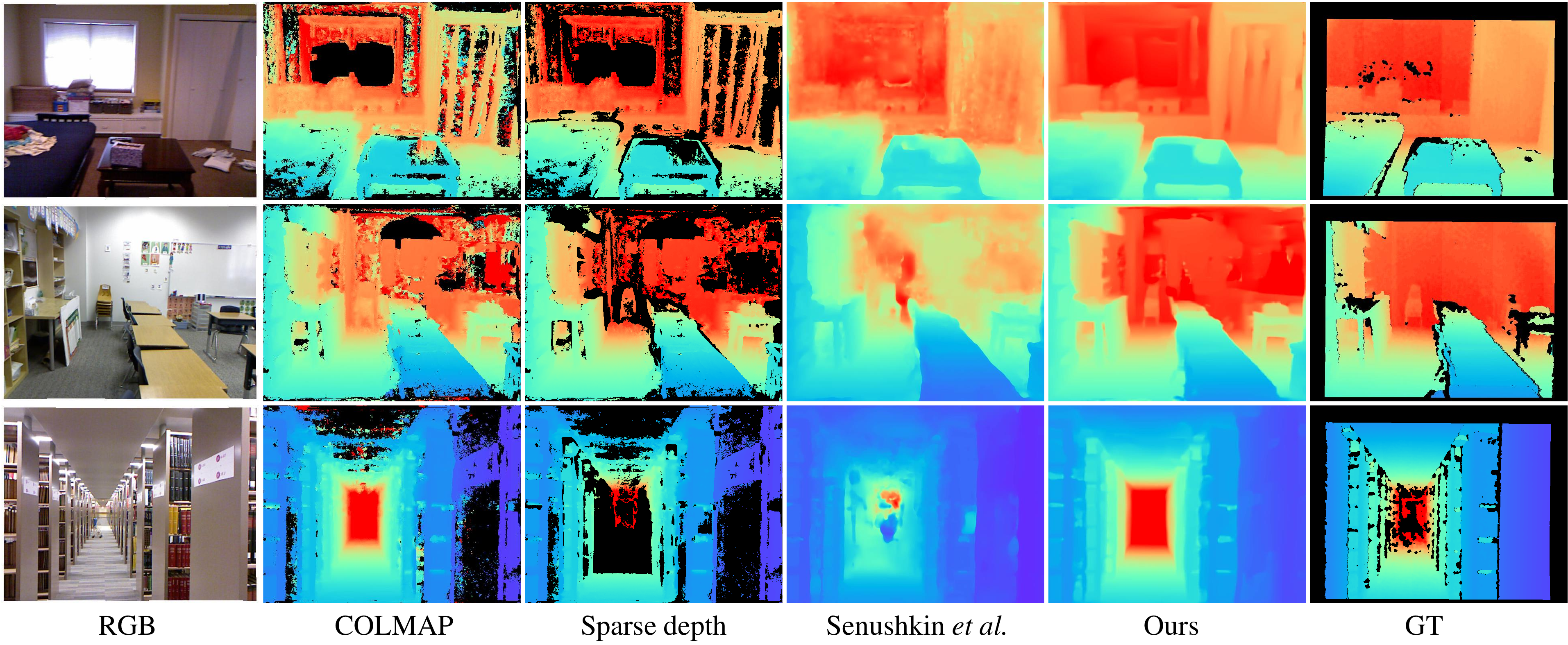}
%\vspace{-0.5 em}
\caption{Qualitative comparison for completing noisy sparse depth. The noisy sparse depths are obtained by masking COLMAP~\cite{schoenberger2016mvs} depths. Our completed results have less outliers and errors.\label{Fig: complete noisy depth visually.}}
\vspace{-1em}
\end{figure*}
% Visual comparison of completing different sparsity patterns
\begin{figure*}
\centering
\includegraphics[width=1\linewidth]{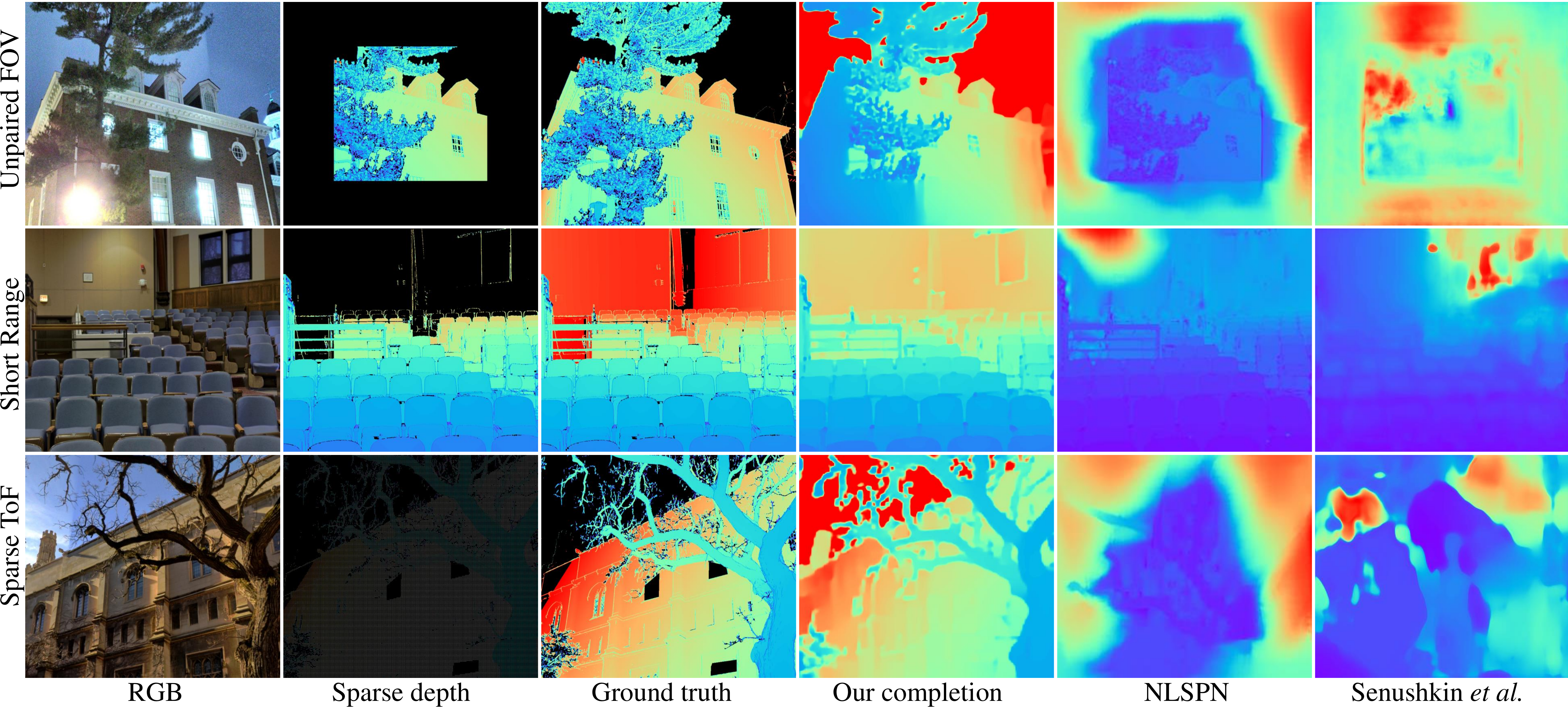}
\vspace{-1.8 em}
\caption{Qualitative completion results on the DIODE~\cite{vasiljevic2019diode} dataset. Note that none of the methods are trained on this dataset. We compare our method with Senushkin~\etal~\cite{senushkin2020decoder} and NLSP~\cite{park2020non} using $3$ different unseen sparsity patterns.
\label{Fig: diff sparsity cmp}}
\vspace{-1 em}
\end{figure*}

\begin{figure*}[t]
\centering
\includegraphics[width=.807\linewidth]{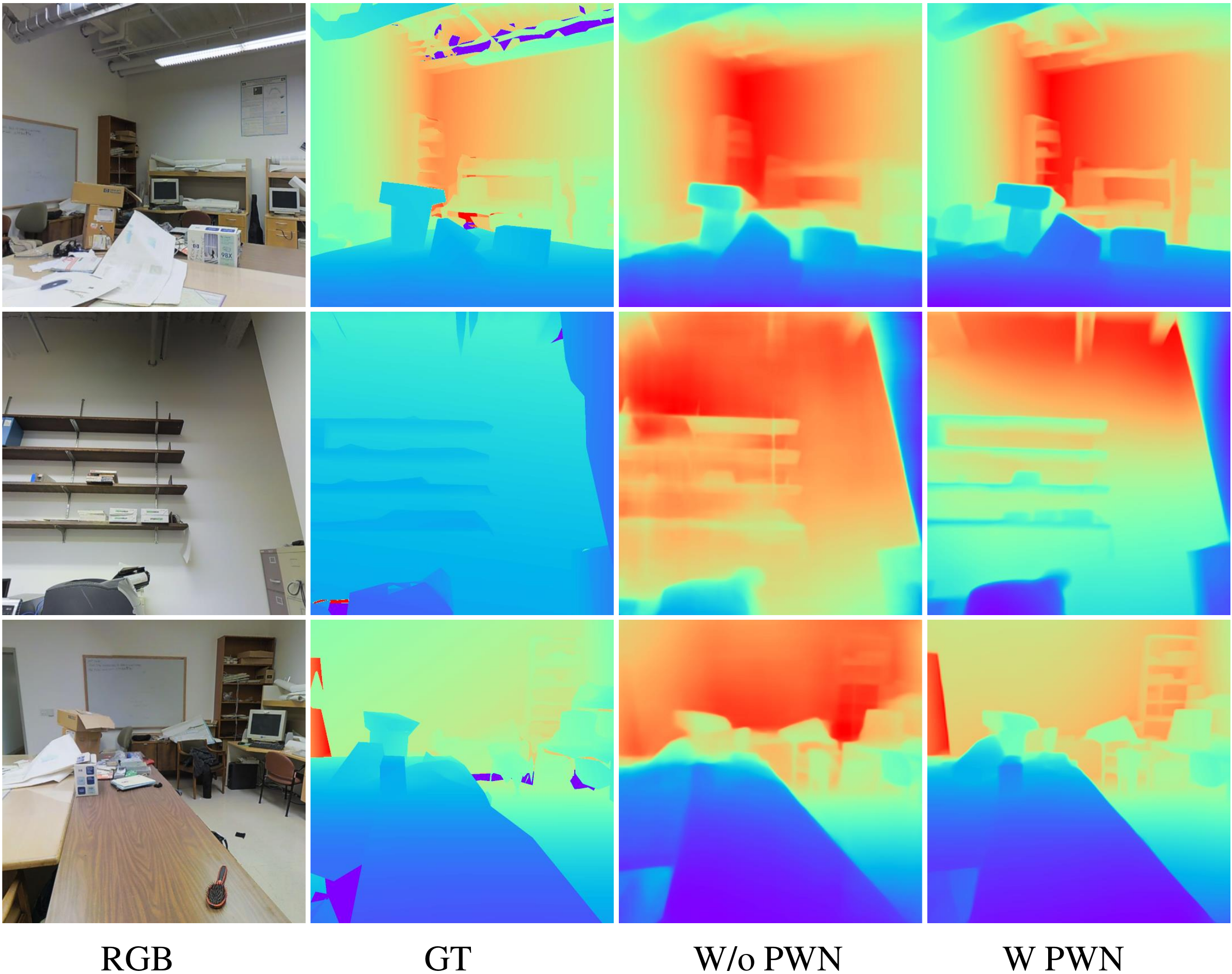}
\caption{\textbf{Qualitative comparison}.  
Using the pair-wise normal loss (PWN), we can see that predicted depths %have
exhibit 
finer details on edges.}
\label{Fig: cmp of depth with PWN. }
%\vspace{-1em}
\end{figure*}

\begin{figure*}[h!]
\centering
\includegraphics[width=.708\linewidth]{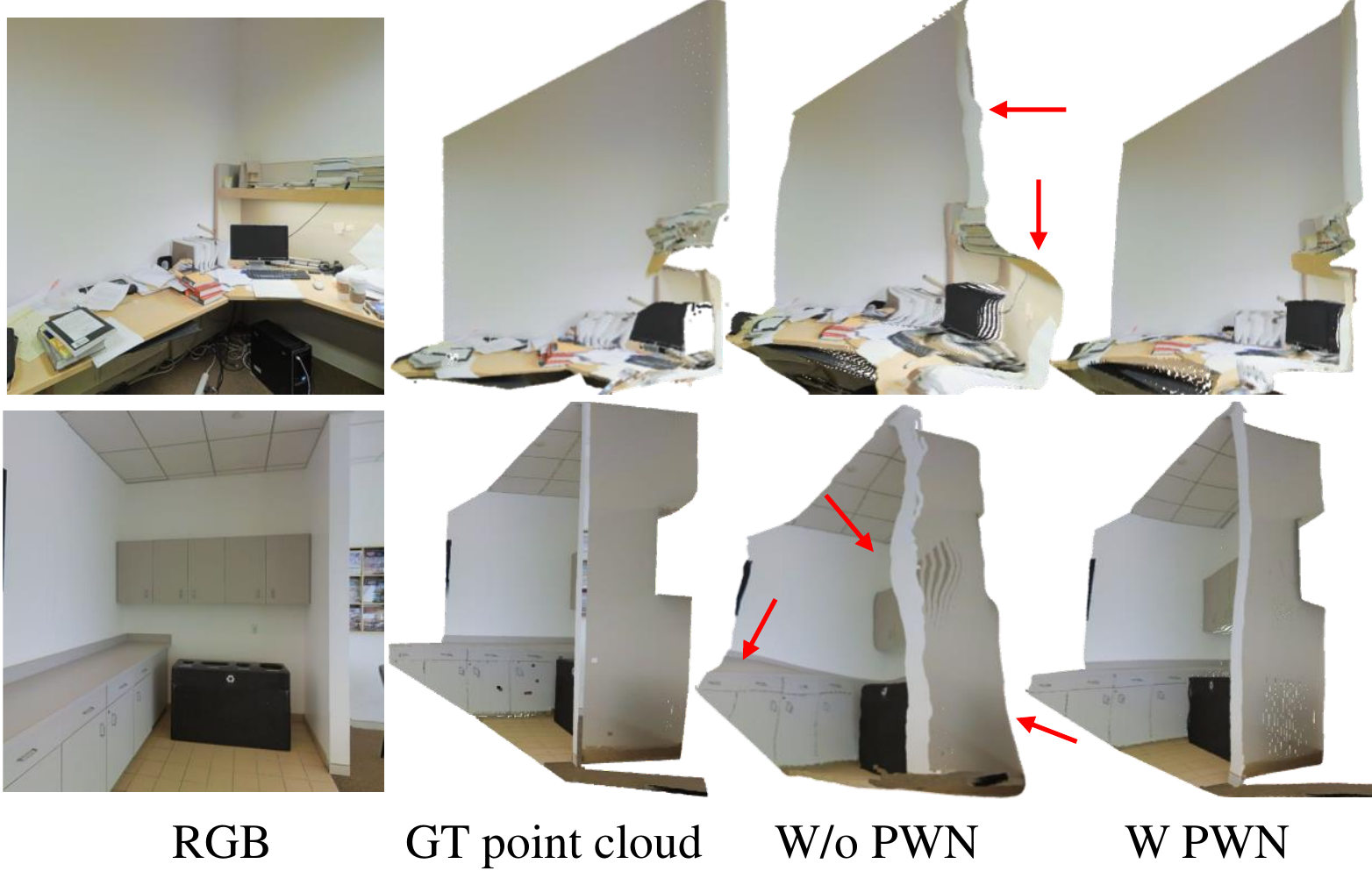}
\caption{\textbf{Qualitative comparison of reconstructed point clouds.} Using the pair-wise normal loss (PWN), we can see that edges and planes are better reconstructed (see highlighted regions).}
\label{Fig: cmp of pcd with PWN. }
%\vspace{-1em}
\end{figure*}

\begin{figure*}[h!]
\centering
\includegraphics[width=.87\linewidth]{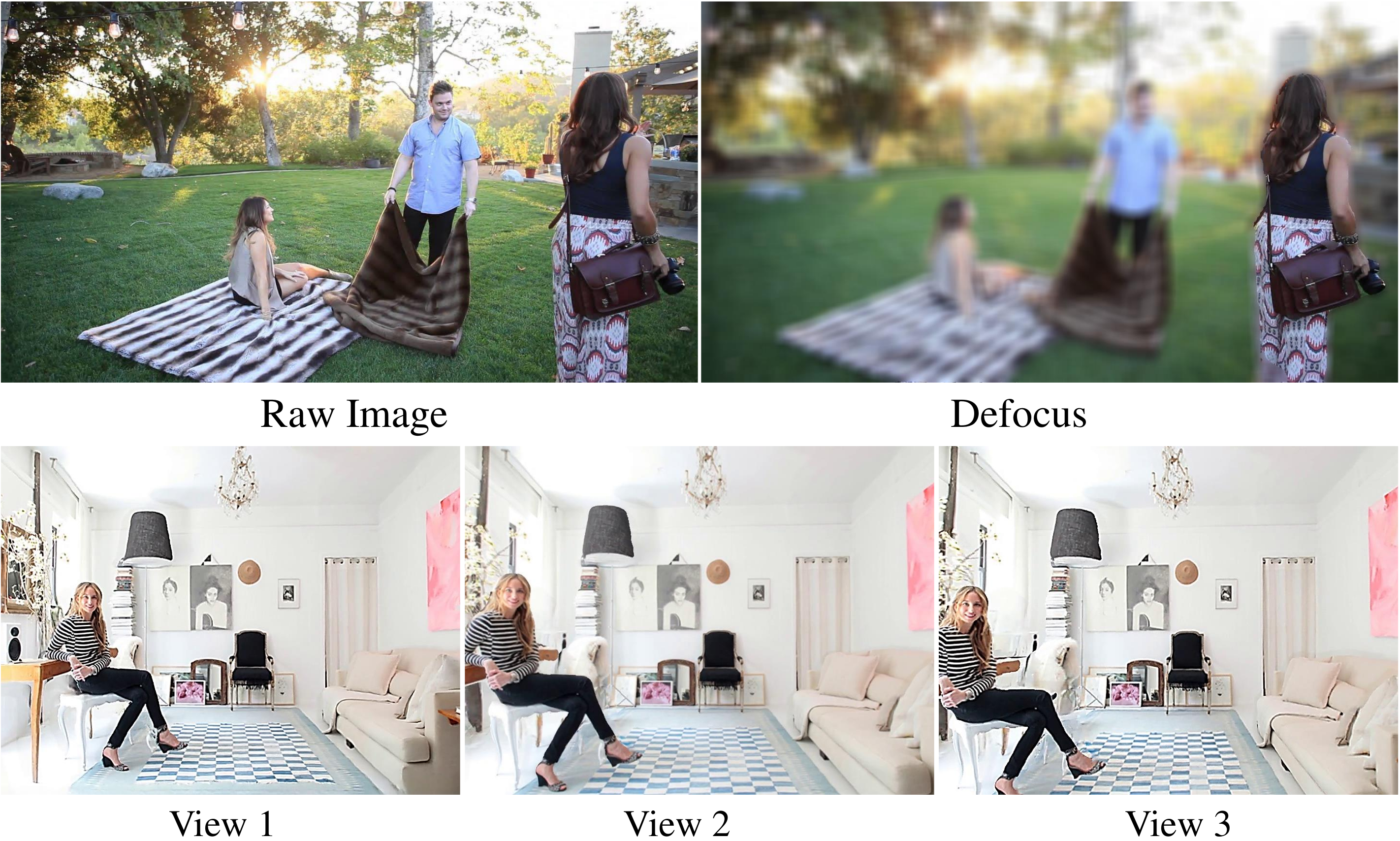}
\caption{
% Qualitative 
\textbf{%Illustration
Demonstration 
of some applications}. We %employ our
use the 
predicted depth to create some visual effects on a single input image. The first row %is defocused,
out of focus synthesis, 
while the second row is new views synthesis.}
\label{Fig: cmp of 3D photo. }
% \vspace{-1 em}
\end{figure*}

\end{document}